\documentclass[runningheads]{llncs}
\usepackage[dvipsnames]{xcolor}
\usepackage{graphicx}
\usepackage{amsmath}
\usepackage{makecell}
\usepackage[caption=false,labelformat=empty]{subfig}

\usepackage{array}
\usepackage{booktabs}
\usepackage{comment}
\usepackage{hyperref}

\usepackage[title]{appendix}
\begin{document}
%
%\title{AI-driven Artifact Monitoring for Cultural Heritage Preservation\thanks{Supported by organisation x. Removed for anonymity}}
%\title{AI-driven Monitoring based on Semantic Segmentation of Artifacts for Cultural Heritage Preservation\thanks{Supported by organisation x. Removed for anonymity}}
%\title{Semantic Segmentation for Automated Monitoring of Cultural Heritage Artifacts}
%\title{Automated Cultural Heritage Artifact Monitoring via Semantic Segmentation}
\title{Automated Monitoring of Cultural Heritage Artifacts Using Semantic Segmentation}

%\author{First Author\inst{1}\orcidID{0000-1111-2222-3333} \and
%Second Author\inst{2,3}\orcidID{1111-2222-3333-4444} \and
%Third Author\inst{3}\orcidID{2222--3333-4444-5555}}
%
\author{A. Ranieri, G. Palmieri, S. Biasotti}
\authorrunning{A. Ranieri et al.}
\begin{comment}
\institute{CNR-IMATI, Genoa, ITALY 
\email{\{andrea.ranieri,giorgio.palmieri,silviamaria.biasotti\}@cnr.it}\\
\url{https://www.imati.cnr.it/} %\and
}
\end{comment}
\institute{CNR-IMATI, Via De Marini, 6 - 16149 Genova (GE), ITALY \email{\{andrea.ranieri,giorgio.palmieri,silviamaria.biasotti\}@cnr.it}}

%
%\institute{Affiliation removed for anonymity}
\maketitle              % typeset the header of the contribution
\begin{abstract}
This paper addresses the critical need for automated crack detection in the preservation of cultural heritage through semantic segmentation. We present a comparative study of U-Net architectures, using various convolutional neural network (CNN) encoders, for pixel-level crack identification on statues and monuments. 
A comparative quantitative evaluation is performed on the test set of the OmniCrack30k dataset~\cite{benz2024omnicrack30k} using popular segmentation metrics including Mean Intersection over Union (mIoU), Dice coefficient, and Jaccard index. This is complemented by an out-of-distribution qualitative evaluation on an unlabeled test set of real-world cracked statues and monuments. 
Our findings provide valuable insights into the capabilities of different CNN-based encoders for fine-grained crack segmentation. We show that the models exhibit promising generalization capabilities to unseen cultural heritage contexts, despite never having been explicitly trained on images of statues or monuments.
%, acknowledging the current lack of public datasets specifically for cultural heritage crack segmentation. 
%Our work utilizes the OmniCrack30k dataset, a large-scale benchmark for universal crack segmentation, both for model training and quantitative evaluation. We train four U-Net models each with a distinct CNN backbone: ResNet-50 (trained with images at 270px resolution), ResNet-101 (at 540px), ConvNeXt V2~\cite{woo2023convnext} Base (at 384px), and ConvNeXt V2 Huge (at 512px). 
%This paper addresses automated crack detection in the preservation of cultural heritage, with a focus on semantic segmentation techniques. Specifically, it presents a comparative study of two leading deep learning approaches—U-Nets and SegFormers—for pixel-level identification of cracks on statues and monuments. To train and evaluate these models, an enhanced and balanced version of the Pothole Mix dataset was adapted for the cultural heritage domain.
%The study offers insights into the comparative performance of convolutional neural network (CNN)-based encoder-decoder models versus Transformer-based architectures in addressing the complex visual challenges posed by heritage artifacts. 
%It highlights the strengths and weaknesses of each approach in achieving accurate and detailed segmentation, guiding future applications of AI in the conservation of cultural heritage.

%\keywords{Deep learning  \and Segmentation \and Monitoring \and Convolutional Neural Networks}
\end{abstract}
\section{Introduction}
\label{sec:introduction}
%AI-driven non-destructive testing (NDT) is revolutionizing cultural heritage preservation by enabling proactive and precise detection of deterioration, moving beyond traditional reactive conservation approaches. This shift is crucial given the increasing threats from natural disasters, climate change, conflicts, and daily degradation.
%AI, particularly deep learning, offers objectivity and scalability, allowing for the identification of subtle deterioration signs that are challenging to detect manually
%AI techniques, and particularly deep learning, offer what we can call "scalabile congnitive automation", allowing for the identification of subtle deterioration signs that are challenging to detect manually, thereby optimizing restoration and protection strategies. Manual inspections are often lengthy, costly, and prone to human error. AI and computer vision techniques provide continuous monitoring and digital records of damage propagation, transforming diagnostics into a standardized, scalable, and data-driven system vital for managing extensive collections.

%This study specifically investigates the use of neural networks for crack detection in cultural heritage, with a focus on marble statues. It reviews existing image datasets for crack segmentation and recent AI models, outlining the current state and future prospects in this specialized field. Detecting cracks on randomly textured surfaces like marble presents unique challenges and has been less explored compared to other materials such as concrete 

The preservation of cultural heritage, encompassing historical statues and monuments, is paramount for understanding human history and artistic achievement. These invaluable artifacts are constantly exposed to environmental degradation, leading to structural damage such as cracks. Early and accurate detection of these cracks is crucial for timely intervention, preventing further deterioration, and ensuring their longevity. Traditional manual inspection methods are often labor-intensive, time-consuming, subjective, and can be limited by accessibility, particularly for large or complex structures. This necessitates the development of automated, efficient, and precise diagnostic tools.

Semantic segmentation, a fundamental task in computer vision, offers a powerful solution by enabling pixel-level classification of images. Unlike object detection, which provides bounding box localization, semantic segmentation precisely delineates the boundaries of objects or regions of interest, making it ideal for identifying fine-grained, irregular structures like cracks. Recent advancements in deep learning, particularly with Convolutional Neural Networks (CNNs), have revolutionized image segmentation, demonstrating superior performance in capturing intricate spatial patterns and semantic representations~\cite{krichen2023convolutional}.

A significant challenge in applying deep learning to cultural heritage preservation is the inherent domain shift and scarcity of annotated data. While large-scale datasets exist for crack detection in civil infrastructure, such as roads and buildings~\cite{benz2024omnicrack30k,thompson2022shrec,elhariri2022historical}, these often present vastly different visual characteristics and crack morphologies compared to the intricate surfaces of statues and monuments. Road cracks, for instance, typically appear on asphalt or concrete, following linear or alligator patterns, whereas cracks in cultural heritage objects can be more subtle, follow material grain (e.g., stone, marble), or occur on metallic surfaces, frequently against complex, textured backgrounds. This domain gap, coupled with the practical impossibility of acquiring large, pixel-level annotated ground truth datasets for unique historical artifacts, mandates a specialized approach to data preparation and evaluation.  

This paper explores the application of state-of-the-art deep learning architectures for semantic segmentation of cracks~\cite{thompson2022shrec,fan2019road} at the domain of cultural heritage. Specifically, we conduct a comparative analysis of U-Nets~\cite{ronneberger2015u}, a widely adopted encoder-decoder architecture known for its efficacy in image segmentation and fine-grained detail preservation, when paired with various CNN backbones as well as in image generation (such as in diffusion models). We utilize the OmniCrack30k dataset~\cite{benz2024omnicrack30k}, a comprehensive benchmark dataset for crack segmentation, for training and quantitative evaluation. Our experimental setup involves training U-Nets with ResNet-50, ResNet-101, ConvNeXt V2 Base, and ConvNeXt V2 Huge as encoders~\cite{he2016deep,woo2023convnext}. Given the scarcity of large, annotated datasets specifically for cultural heritage crack detection, we employ a two-fold evaluation approach: a quantitative assessment on the OmniCrack30k test set and a qualitative evaluation on an unlabeled test set of real-world cracked statues and monuments.\\ 

The main contributions of this work are:
\begin{itemize}
\item An analysis of the applicability and performance of U-Net architectures with diverse CNN encoders for semantic segmentation of cracks on cultural heritage artifacts, including different CNN backbone complexities.
%\item An exploration of different CNN backbone complexities for U-Net, assessing their impact on crack detection quality.
\item A detailed methodology for leveraging the OmniCrack30k dataset for training, addressing the challenges of data diversity in crack segmentation.
\item A comparative quantitative evaluation of model performance on the OmniCrack30k test set using mIoU, Dice, and Jaccard metrics.
\item A qualitative evaluation framework for assessing model performance on an unlabeled test set of real-world cracked statues and monuments, providing practical insights for conservators and highlighting generalization capabilities to unseen domains.
\item A \textbf{public repository containing all the code}\footnote{\url{https://gitlab.com/4ndr3aR/cultural-artifacts-crack-segmentation}} used to produce (and therefore to replicate) this work and the pre-trained models used to generate the images in this paper.
\end{itemize}

\section{Related Work}
Automated crack segmentation is a critical component in structural health monitoring and integrity systems, with applications spanning road infrastructure, buildings, and cultural heritage. Research in this field has evolved from traditional image processing techniques to sophisticated deep learning paradigms.  

\subsection{Traditional vs. Deep Learning Approaches for Crack Segmentation}
While these methods are computationally less intensive, traditional computer vision approaches often struggle with complex crack image backgrounds, varying illumination, noise, and the intricate spatial details of cracks, leading to limited effectiveness and reliability in heterogeneous or noisy environments.  
Early methods for crack detection relied on traditional image processing techniques such as edge detection (e.g., Canny~\cite{landstrom2012morphology,zhou2021mixed}, Sobel~\cite{dixit2018investigating}), thresholding~\cite{sari2019road}, morphological operations~\cite{zhen2025crack}, and statistical analysis.%~\cite{kumar2013algorithm,munawar2021image}

The advent of deep learning has significantly advanced crack segmentation. Deep learning approaches, particularly Convolutional Neural Networks (CNNs) can directly learn complex spatial patterns and feature representations from labeled data. Their ability to extract and match the most relevant features significantly improves the segmentation accuracy.  

\subsection{CNN and UNet-based Architectures}
\label{sec:star-architectures}
%\textcolor{blue}{
In the domain of cultural heritage, the scientific community has primarily focused on the classification and the object detection of cracks.
In \cite{chaiyasarn2018crack}, the authors construct a dataset of 6002 images for binary crack classification on images of temples captured by an AUV in the Ayutthaya region of Thailand. They then train a three-layer CNN by comparing three different types of classifiers for CNN features: the fully connected layer of the CNN itself, a Support Vector Machine (SVM), and a Random Forest (RF).
In \cite{karimi2025automated}, the researchers propose a dataset of 4,374 images for object detection of cracks on various masonry materials (cob, brick, stone, and tile). The authors share their dataset publicly on Kaggle and train a YOLOv5 model on it, achieving mAP50 values between 94.4\% and 70.3\%, depending on the material.%}

U-Net, introduced by Ronneberger et al. \cite{u-net}, is a seminal deep learning architecture widely adopted for image segmentation. %, particularly in biomedical imaging due to its ability to handle limited annotated data while maintaining speed and accuracy. 
Its "U"-shaped architecture comprises a contracting path (encoder) that progressively down-samples the input image through convolutional layers and pooling operations, capturing contextual information and reducing spatial resolution. The expansive path (decoder) then up-samples these features, gradually restoring spatial resolution and generating a segmentation map. A key innovation of U-Net is to directly concatenate feature maps from the contracting path to the corresponding decoder layers. This mechanism is crucial for preserving fine-grained spatial information that might otherwise be lost during downsampling, enabling the network to produce high-resolution predictions with precise boundaries.  

%Variants of U-Net and other Fully Convolutional Networks (FCNs)  have been extensively used for crack detection. FCN architectures employ skip connections to integrate features across multiple network layers, allowing a seamless fusion of high-level semantic information with low-level spatial details during upsampling.  

For the U-Net models, we utilize four prominent CNN backbones as encoders:
\begin{itemize}
\item \textbf{ResNet-50:} ResNet-50 is a 50-layer deep convolutional neural network known for addressing the vanishing gradient problem through residual blocks and shortcut connections. %It is widely used as a feature extractor in various computer vision tasks, often pre-trained on large datasets like ImageNet.
We use a standard \textit{Torchvision} model with 25.6 M parameters, fine-tuned on images resized to 270x270 pixels.
\item \textbf{ResNet-101:} ResNet-101 is a deeper variant of ResNet-50, comprising 101 layers (44.5 M parameters). 
%Its increased depth allows it to extract more advanced and abstract features compared to ResNet-50, potentially leading to improved segmentation accuracy. 
This model is trained on images resized to 540x540 pixels, leveraging its greater capacity for capturing finer details at a larger scale.
\item \textbf{ConvNeXt V2 Base:} ConvNeXt V2 is a purely convolutional architecture that significantly improves the performance of ConvNets, particularly when co-designed with masked autoencoders for self-supervised learning. 
It is pretrained on the popular \textit{ImageNet-22k} dataset and is known for its effective design, balancing model complexity (88.7 M parameters) and performance. %ConvNeXt architectures have been successfully integrated as encoders in U-Net-like structures for semantic segmentation, demonstrating strong feature extraction capabilities and computational efficiency. 
We fine-tune this model on images resized to 384x384 pixels.
%The ConvNeXt V2 Base model has approximately 88.7 million parameters.
\item \textbf{ConvNeXt V2 Huge:} This is the largest variant of the ConvNeXt V2 family (660 M parameters). % achieves state-of-the-art performance using only public training data. Its substantial capacity allows for the extraction of highly robust and abstract features, making it suitable for complex segmentation tasks.
Like its Base counterpart, it is pre-trained on the \textit{ImageNet-22k} dataset and has been fine-tuned using images resized to 512x512 pixels to maximize detail capture and segmentation precision.
%, boasting approximately 650 million parameters
\end{itemize}
The decoder for all U-Net variants follows the standard U-Net design, incorporating upsampling layers and concatenating features from the respective CNN encoder via skip connections.  

\subsection{Crack Detection in Cultural Heritage and Dataset Challenges}
While crack detection is widely studied for civil infrastructures \cite{thompson2022shrec,fan2019road}, its application to CH poses unique challenges due to its peculiarities. % due to the diverse materials, intricate geometries, and often unique, irreplaceable nature of the artifacts. 
Existing datasets for cultural heritage are scarce and often small, making large-scale supervised training difficult. For instance, the Historical-Crack18-19 dataset \cite{elhariri2022historical} contains 3886 annotated images from an ancient mosque, but only 757 are crack images, highlighting the imbalance and limited scope. %Studies have explored CNNs for monument recognition  and U-Net for crack area segmentation in historical buildings, concluding U-Net's suitability for segmentation.  

The scarcity of high-quality, labeled vision data for damaged artifacts is a significant impediment to developing robust deep learning models. This often necessitates data augmentation strategies, including traditional methods like rotations, resizing, and noise addition, or more advanced techniques like synthetic data generation. Synthetic data can augment real-world datasets, improving model performance and generalization, especially for rare defect occurrences. Tools like \textit{UnrealROX} \cite{martinez2020unrealrox} and \textit{EasySynth} \cite{Jovanovic2025} can generate realistic synthetic images to overcome training data limitations.
%CrackMover is a novel data augmentation strategy specifically tailored for crack detection by applying random transformations to existing mask annotations to generate new images, thereby increasing data variability and model robustness. These data adaptation and augmentation techniques are critical methodological contributions that enable the effective transfer of knowledge from a source domain (e.g., general crack images) to a target domain (cultural heritage cracks) where labeled data is scarce.

\section{Methodology}
%This section details the problem definition, the dataset utilized, the deep learning architectures employed, the experimental setup, and the protocol for quantitative and qualitative evaluation.

\subsection{Problem Definition and Scope}
Our objective is to perform semantic segmentation of cracks on the surfaces of statues and monuments. This involves assigning a binary label (crack or non-crack) to each pixel in an input image, thereby precisely delineating the extent and morphology of the damage. The inherent challenges in this task include the fine-grained nature of cracks, their variable morphology (e.g., hairline or structural), the complex and often textured backgrounds of cultural heritage artifacts, and the significant scarcity of annotated ground truth data for this specific domain. We also conduct a comparative study of the ability of the U-Net architectures with various CNN backbones to address these challenges.

\subsection{Dataset for Training and Quantitative Evaluation}
The primary dataset for our study is OmniCrack30k. OmniCrack30k is a large-scale, systematic, and thorough benchmark dataset specifically designed for universal crack segmentation. It comprises 30,000 samples compiled from over 20 diverse datasets, for a total of 9 billion pixels. This compilation features images of cracks on a wide array of materials, including asphalt, ceramic, concrete, masonry, and steel. The dataset's comprehensive nature aims to reduce biases and enable robust benchmarking for general crack segmentation tasks.  

While OmniCrack30k provides extensive data on various cracked materials, it does not specifically include images of statues or monuments. This inherent diversity, however, makes it an excellent foundation for training models that can generalize to different surface textures and crack patterns. To enhance model robustness and generalization, standard data augmentation techniques are also applied during training. The dataset is split into training, validation, and test sets, with the test set used only for the quantitative evaluation of our fine-tuned models.

%\subsection{Deep Learning Architectures}
\subsection{Experimental Setup}

We fine-tune and evaluate the U-Net architectures with four distinct CNN backbones, representing a range of model capacities as introduced in Section \ref{sec:star-architectures}.
The training took place within a Jupyter Notebook environment running Python 3.10 and using the popular \textit{Fast.ai} library now at its second version~\cite{fastai}. \textit{Fast.ai} adds an additional layer of abstraction above \textit{Pytorch}~\cite{paszke2019pytorch}, therefore it is very convenient to use for speeding up the "standard" and repetitive tasks of training a neural network.

\subsubsection{Data augmentation}
We train the four architectures both with and without data augmentation. For the data augmentation pipeline, we employ the popular \textit{Albumentations} library. We apply a stochastic augmentation pipeline with probability $p=1$, that combines the following data-augmentation branches:

\begin{itemize}
    \item geometric transformations (\textit{HorizontalFlip}, \textit{RandomRotate90}, \textit{Transpose},\\ \textit{ShiftScaleRotate}) at $p=0.25$ per operation;
    \item moderate distortions (\textit{Blur}, \textit{ElasticTransform}, \textit{GridDistortion}, \textit{OpticalDistortion}) are introduced with probability $p=0.1$ with limited displacement and distortion intensities;
    \item photometric variations via \textit{HueSaturationValue} and \textit{CLAHE} ($p=0.1$ each) simulate illumination and contrast shifts.
\end{itemize}

\subsubsection{Optimization and Hyperparameters}
All models are trained using the Adam optimizer, known for its adaptive learning rate capabilities, initialized with the default \textit{Fast.ai} parameters. A cosine annealing learning rate schedule is employed, starting with an initial learning rate that ranges between $1e^{-3}$ and $1e^{-4}$. The batch sizes are set to 12 and 8 for the two U-Nets with ResNet backbones and to 24 and 5 for the two U-Nets with ConvNext V2 backbones.
For the loss function, we employ the standard \textbf{Binary Cross-Entropy (BCE) Loss} that calculates probabilities and compares each actual class output with predicted ones, making it suitable for pixel-level binary classification (crack vs. background).

\subsubsection{Training details}
To maximize the level of automation during the training of the network, some Fast.ai callbacks have been used to perform the early stopping of the training (with $patience = 2$, i.e. the training stops when the validation loss of the network does not improve for $2$ consecutive epochs) and to automatically save the best model of the current training round (according to both validation loss, Dice and Jaccard metrics). Later, that model is reloaded for the final validation phase and to show predicted images on the validation and test set. Experiments were conducted on a workstation equipped with three Nvidia RTX A6000 GPUs (48 Gb of VRAM each), an AMD Ryzen Threadripper Pro 7965WX CPU (with 24 cores/48 threads) and 128 Gb of DDR5 RAM. Each each model was trained on only one GPU at a time, so training times reported in \ref{tab:train_val_metrics_no_data_aug} refer to single-GPU training runs.
%All the training runs were performed on a workstation equipped with a Ryzen Threadripper 2x Nvidia RTX A6000 GPUs with 48 Gb of VRAM each, but each model was trained on only one GPU at a time.

\subsection{Evaluation Protocol}
Our evaluation protocol consists of two distinct phases to thoroughly assess model performance.

\subsubsection{Quantitative Evaluation on OmniCrack30k Test Set}
Upon completion of training, a comparative quantitative evaluation of the different models is performed on the dedicated test set of the OmniCrack30k dataset. The following popular segmentation metrics are calculated: \textbf{i) Mean Intersection over Union (mIoU)}; \textbf{ii) Dice Coefficient}, and \textbf{iii) Jaccard Index:} \cite{Deza.Deza2009EncyclopediaofDistances}.

%\begin{itemize}
%\item \textbf{Mean Intersection over Union (mIoU):} A standard metric for semantic segmentation, calculated as the average of the IoU scores for each class (crack and background). IoU measures the overlap between the predicted segmentation and the ground truth.
%\item \textbf{Dice Coefficient:} Also known as F1-score, it measures the similarity between two sets (predicted and ground truth pixels). It is particularly robust to class imbalance and is widely used for segmentation tasks.
%\item \textbf{Jaccard Index:} Similar to IoU, it quantifies the similarity and diversity of sample sets.
%\end{itemize}
These metrics provide an objective measure of how well each model performs pixel-wise crack segmentation on a diverse, large-scale dataset.  

\subsubsection{Out-of-Distribution Qualitative Evaluation}

Given the lack of publicly available, annotated datasets specifically for cracked statues and monuments for semantic segmentation, an out-of-distribution qualitative evaluation is performed. This involves applying the trained models to an unlabeled test set comprising real damaged and cracked statues and monuments downloaded from the web. The results are then qualitatively rated by the authors, focusing on: \textbf{i) Crack Continuity and Completeness}, that is, how well the models detect entire crack networks; \textbf{ii) Boundary Precision} accuracy of crack outlines. \textbf{iii) False Positives/Negatives:}, instances of over-segmentation (e.g., misclassifying textures as cracks) or under-segmentation (missing actual cracks); \textbf{iv) Generalization:}, the models' ability to perform on visually distinct materials and environments not seen during training.

%\begin{itemize}
%\item \textbf{Crack Continuity and Completeness:} How well the models detect entire crack networks.
%\item \textbf{Boundary Precision:} Accuracy of crack outlines.
%\item \textbf{False Positives/Negatives:} Instances of over-segmentation (e.g., misclassifying textures as cracks) or under-segmentation (missing actual cracks).
%\item \textbf{Generalization:} The models' ability to perform on visually distinct materials and environments not seen during training.
%\end{itemize}
This qualitative assessment is crucial for understanding the practical utility and generalization capabilities of the models in a real-world cultural heritage context.  

\section{Experiments}
%This section presents the quantitative evaluation of the U-Net models on the OmniCrack30k test set, followed by a qualitative assessment of their performance on out-of-distribution images of cracked statues and monuments.

\vspace{-8mm}           % SALVASPAZIO!!!
\begin{table}[tbh]
\centering
\begin{tabular}{|l|c|c|c|c|c|c|}
\hline
\textbf{U-Net Architecture} & \textbf{Train Loss} & \textbf{Val Loss} & \textbf{mIoU} & \textbf{Dice} & \textbf{Jaccard} & \textbf{\makecell{Time \\\ per \\ Epoch \\ (h:mm)}} \\
\hline
ResNet-50, 270px        & 0.026 & 0.027 & 0.634 & 0.840 & 0.755 & 0:19 \\
ResNet-101, 540px       & 0.031 & 0.030 & 0.624 & 0.821 & 0.735 & 1:06 \\
ConvNeXt V2 Base, 384px & 0.022 & 0.026 & 0.638 & 0.848 & 0.765 & 0:24 \\
ConvNeXt V2 Huge, 512px & 0.021 & 0.025 & 0.641 & 0.859 & 0.778 & 5:33 \\
\hline
\end{tabular}
\caption{Quantitative Metrics (best epoch) on OmniCrack30k Training and Validation Sets (no data augmentation regime)}
\label{tab:train_val_metrics_no_data_aug}
\end{table}

\vspace{-16mm}           % SALVASPAZIO!!!
\begin{table}[tbh]
\centering
\begin{tabular}{|l|c|c|c|c|c|c|}
\hline
\textbf{U-Net Architecture} & \textbf{Train Loss} & \textbf{Val Loss} & \textbf{mIoU} & \textbf{Dice} & \textbf{Jaccard} & \textbf{\makecell{Time \\\ per \\ Epoch \\ (h:mm)}} \\
\hline
ResNet-50, 270px        & 0.027 & 0.026 & 0.626 & 0.828 & 0.742 & 0:20 \\
ResNet-101, 540px       & 0.034 & 0.035 & 0.619 & 0.808 & 0.720 & 1:06 \\
ConvNeXt V2 Base, 384px & 0.033 & 0.030 & 0.626 & 0.822 & 0.736 & 0:24 \\
ConvNeXt V2 Huge, 512px & 0.028 & 0.026 & 0.636 & 0.851 & 0.768 & 5:33 \\
\hline
\end{tabular}
\caption{Quantitative Metrics (best epoch) on OmniCrack30k Training and  Validation Sets (with data augmentation regime)}
\label{tab:train_val_metrics_basic_data_aug}
\end{table}

\vspace{-4mm}           % SALVASPAZIO!!!
\begin{table}[tbh]
\centering
\begin{tabular}{|l|c|c|c|c|}
\hline
\textbf{U-Net Architecture} & \textbf{Test Loss} & \textbf{mIoU} & \textbf{Dice} & \textbf{Jaccard} \\
\hline
ResNet-50, 270px        & 0.024 & 0.652 & 0.853 & 0.771 \\
ResNet-101, 540px       & 0.029 & 0.645 & 0.831 & 0.745 \\
ConvNeXt V2 Base, 384px & 0.026 & 0.662 & 0.852 & 0.770 \\
ConvNeXt V2 Huge, 512px & 0.024 & 0.666 & 0.865 & 0.786 \\
\hline
\end{tabular}
\caption{Quantitative Metrics (best epoch) on OmniCrack30k Test Set (no data augmentation regime)}
\label{tab:test_metrics_no_data_aug}
\end{table}

\vspace{-2mm}           % SALVASPAZIO!!!
\begin{table}[tbh]
\centering
\begin{tabular}{|l|c|c|c|c|}
\hline
\textbf{U-Net Architecture} & \textbf{Test Loss} & \textbf{mIoU} & \textbf{Dice} & \textbf{Jaccard} \\
\hline
ResNet-50, 270px        & 0.025 & 0.651 & 0.841 & 0.757 \\
ResNet-101, 540px       & 0.034 & 0.645 & 0.822 & 0.734 \\
ConvNeXt V2 Base, 384px & 0.030 & 0.647 & 0.830 & 0.744 \\
ConvNeXt V2 Huge, 512px & 0.024 & 0.659 & 0.862 & 0.782 \\
\hline
\end{tabular}
\caption{Quantitative Metrics (best epoch) on OmniCrack30k Test Set (with data augmentation regime)}
\label{tab:test_metrics_basic_data_aug}
\end{table}
\vspace{-2mm}           % SALVASPAZIO!!!

%There is a clear trend of improved performance with increasing model complexity and input resolution. The U-Net with ConvNeXt V2 Huge backbone, trained at 512px, consistently achieves the lowest loss and highest mIoU, Dice, and Jaccard scores. This indicates its superior ability to accurately segment cracks within the diverse OmniCrack30k dataset. The ConvNeXt V2 Base model also outperforms the ResNet-based U-Nets, suggesting the benefits of its architecture and masked autoencoder pre-training. The ResNet-101 U-Net shows an improvement over ResNet-50, highlighting the advantage of a deeper backbone and higher resolution for capturing more intricate features. The training time per epoch generally increases with model size and input resolution, reflecting the higher computational demands.

\begin{figure}[!hb]
\centering
\subfloat[\label{fig:ConvNext-Huge-1-predictions-on-material_a}]{\includegraphics[trim={0 0 223px 0},clip,width=0.49\textwidth]{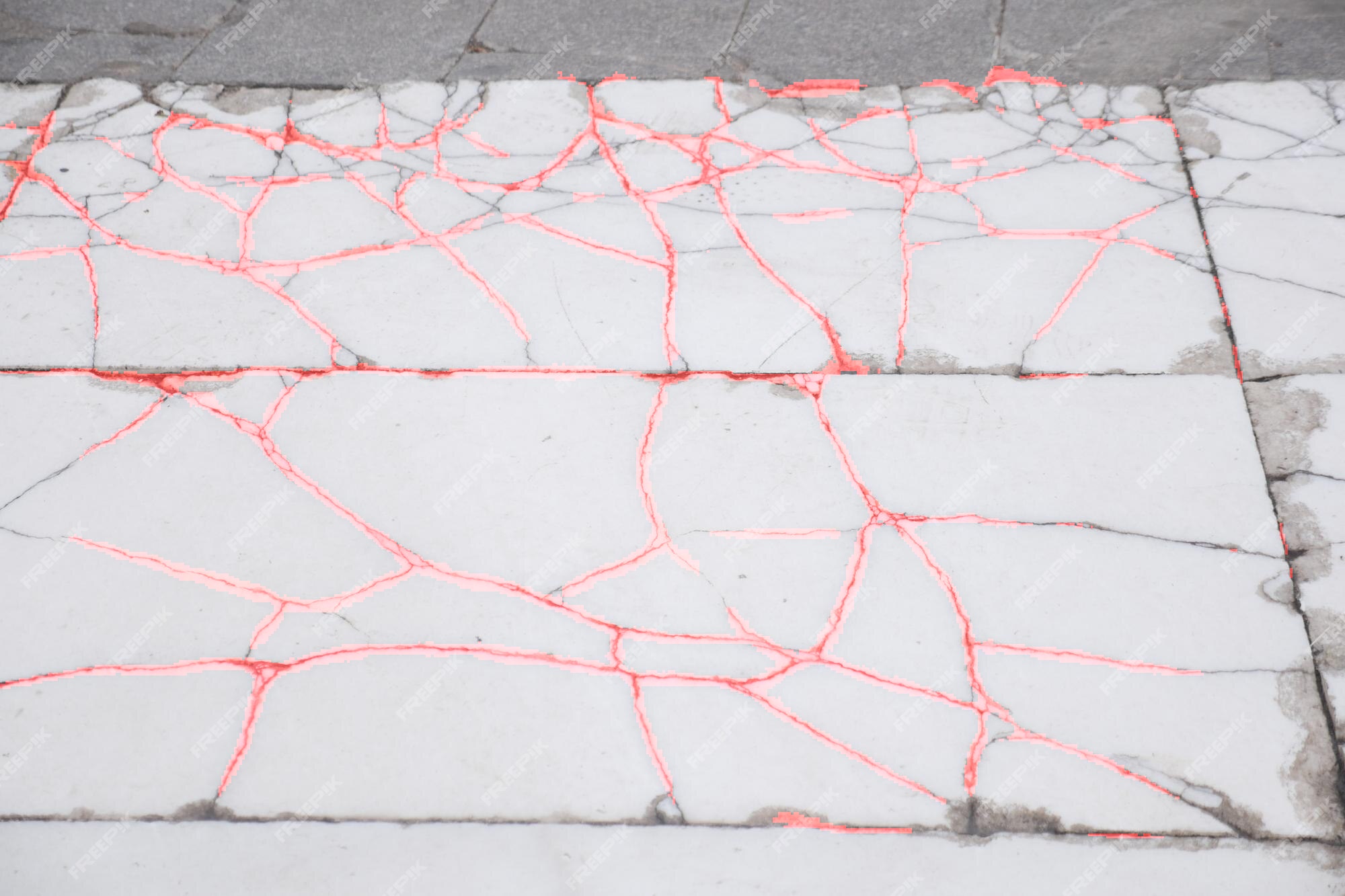}}
\hfil
\subfloat[\label{fig:ConvNext-Huge-1-predictions-on-material_b}]{\includegraphics[trim={0 0 0 0},clip,width=0.49\textwidth]{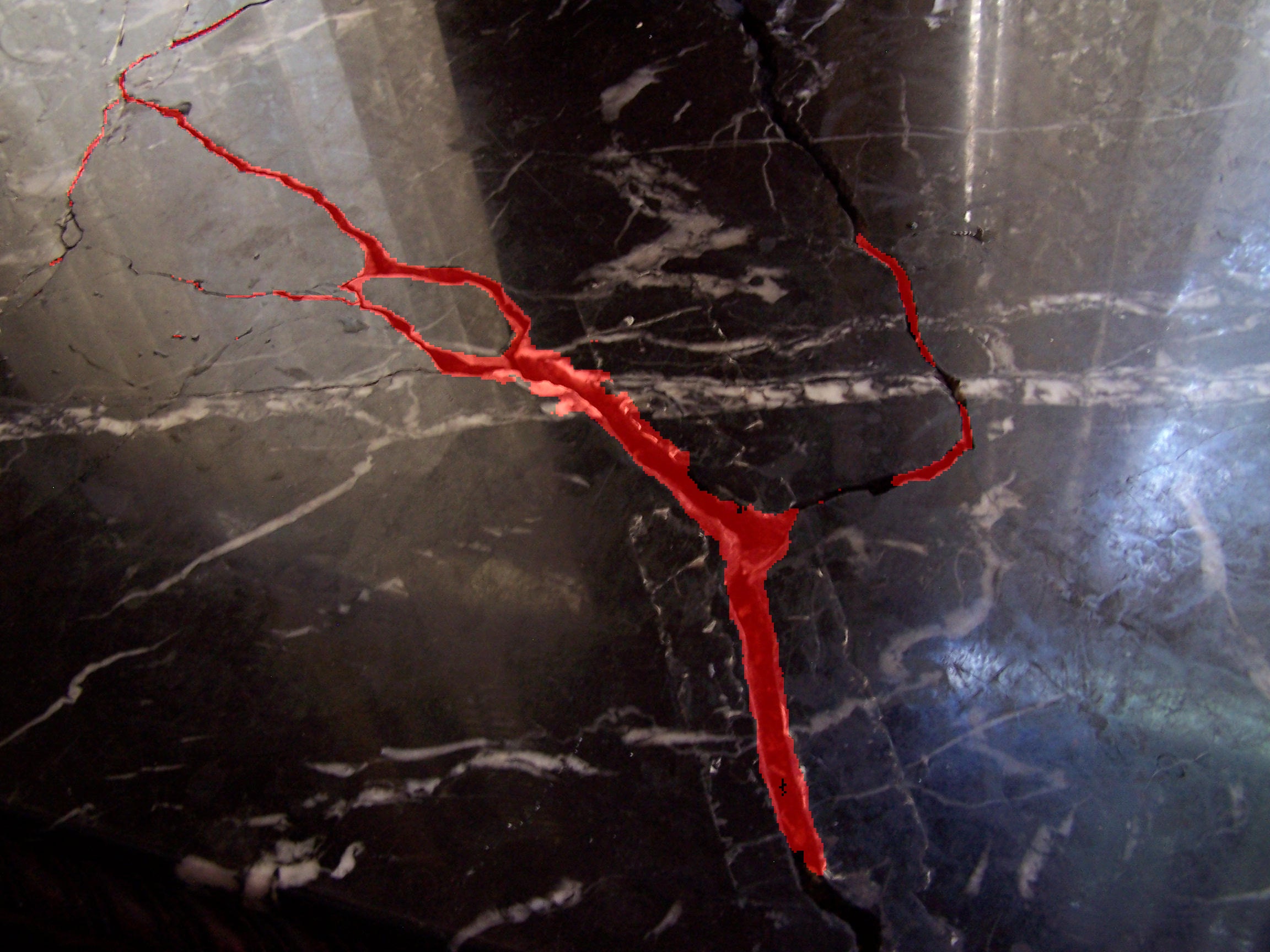}}
\vspace{-6mm}           % SALVASPAZIO!!!
\caption{Out-of-distribution predictions obtained with the \textit{ConvNeXt V2 Huge} U-Net model (no data augmentation regime) on images of black and white marble (therefore closer to images in the training set).}
\label{fig:ConvNext-Huge-1-predictions-on-material}
\end{figure}

\begin{comment}
\begin{figure}[hb]
    \centering
    % Prima immagine
    \begin{subfigure}[b]{0.45\textwidth}
        \centering
        \includegraphics[trim={0 0 223px 0},clip,width=\textwidth]{pics/inference/convnextv2_huge-NO-data-aug/Image_18-blend-512-512.jpg}
        %\caption{}
        %\label{subfig:ConvNext-Huge-18}
    \end{subfigure}
    %\hfill % Spazio orizzontale tra le immagini
    % Seconda immagine
    \begin{subfigure}[b]{0.45\textwidth}
        \centering
        \includegraphics[trim={0 0 0 0},clip,width=\textwidth]{pics/inference/convnextv2_huge-NO-data-aug/Image_58-blend-512-512.jpg}
        %\caption{}
        %\label{subfig:ConvNext-Huge-58}
    \end{subfigure}
    %\caption{Images generated with the \textit{ConvNeXt V2 Huge} U-Net model (no data augmentation regime).}
    \caption{Out-of-distribution predictions obtained with the \textit{ConvNeXt V2 Huge} U-Net model (no data augmentation regime) on images of black and white marble (therefore closer to images in the training set).}
    \label{fig:ConvNext-Huge-1-predictions-on-material}
\end{figure}
%\vspace{-2mm}           % SALVASPAZIO!!!
\end{comment}

\begin{figure}[!ht]
\centering
\subfloat[(a) ResNet-50\label{fig:other-predictions-on-material_a}]{\includegraphics[trim={0 0 0 0},clip,width=0.325\textwidth]{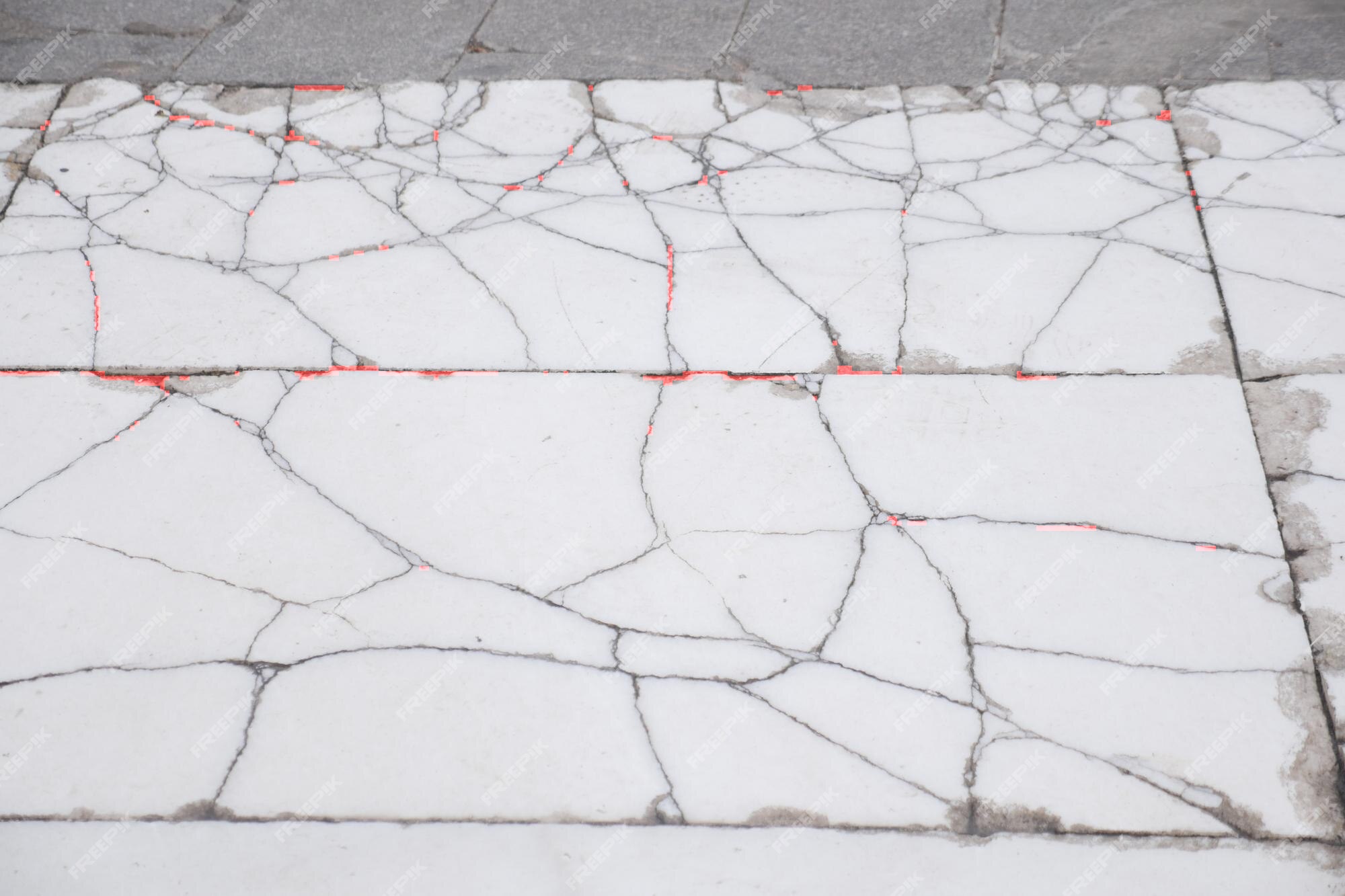}}
\hfil
\subfloat[(b) ResNet-101\label{fig:other-predictions-on-material_b}]{\includegraphics[trim={0 0 0 0},clip,width=0.325\textwidth]{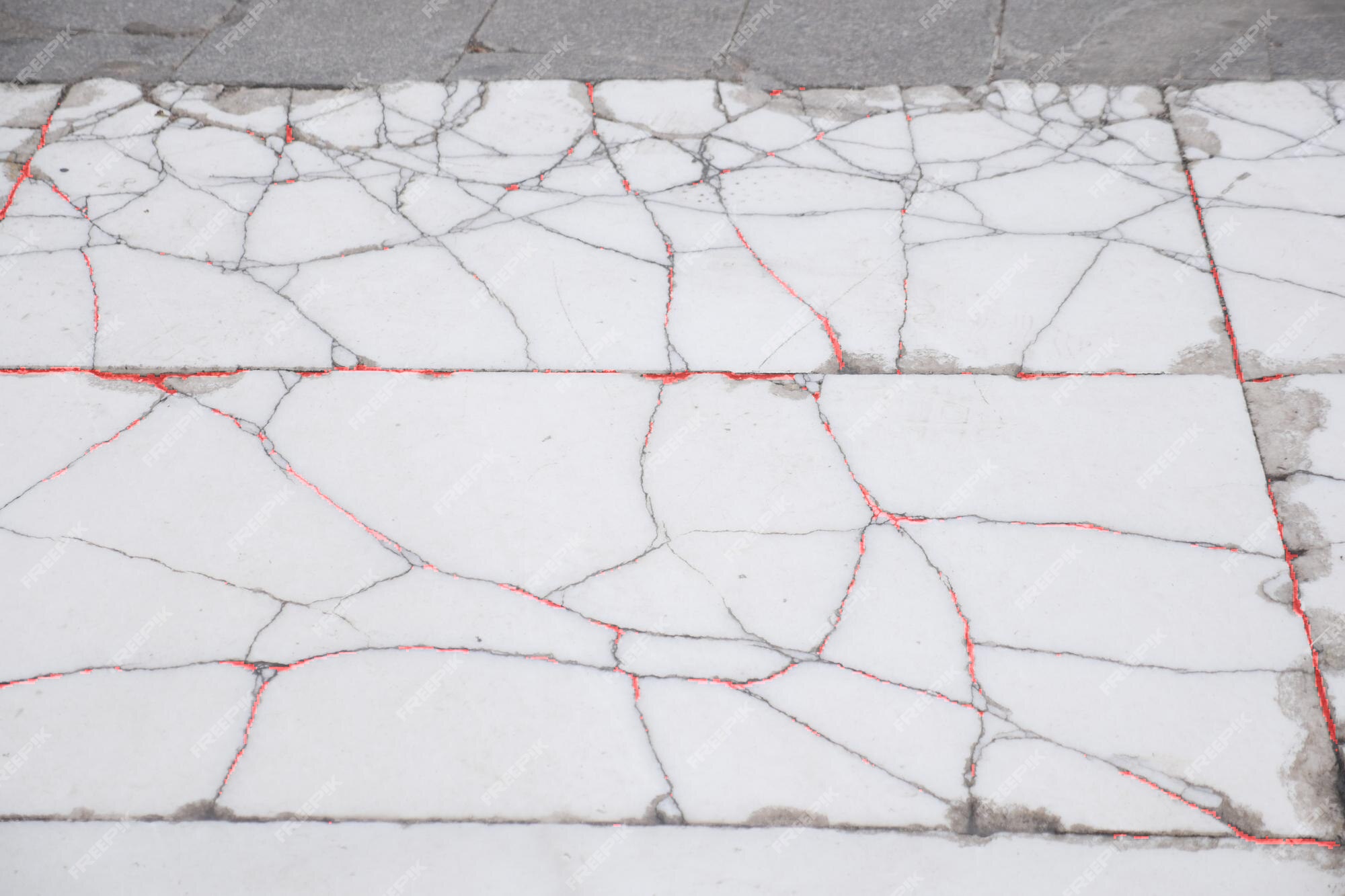}}
\hfil
\subfloat[(c) ConvNeXt V2 Base\label{fig:other-predictions-on-material_c}]{\includegraphics[trim={0 0 0 0},clip,width=0.325\textwidth]{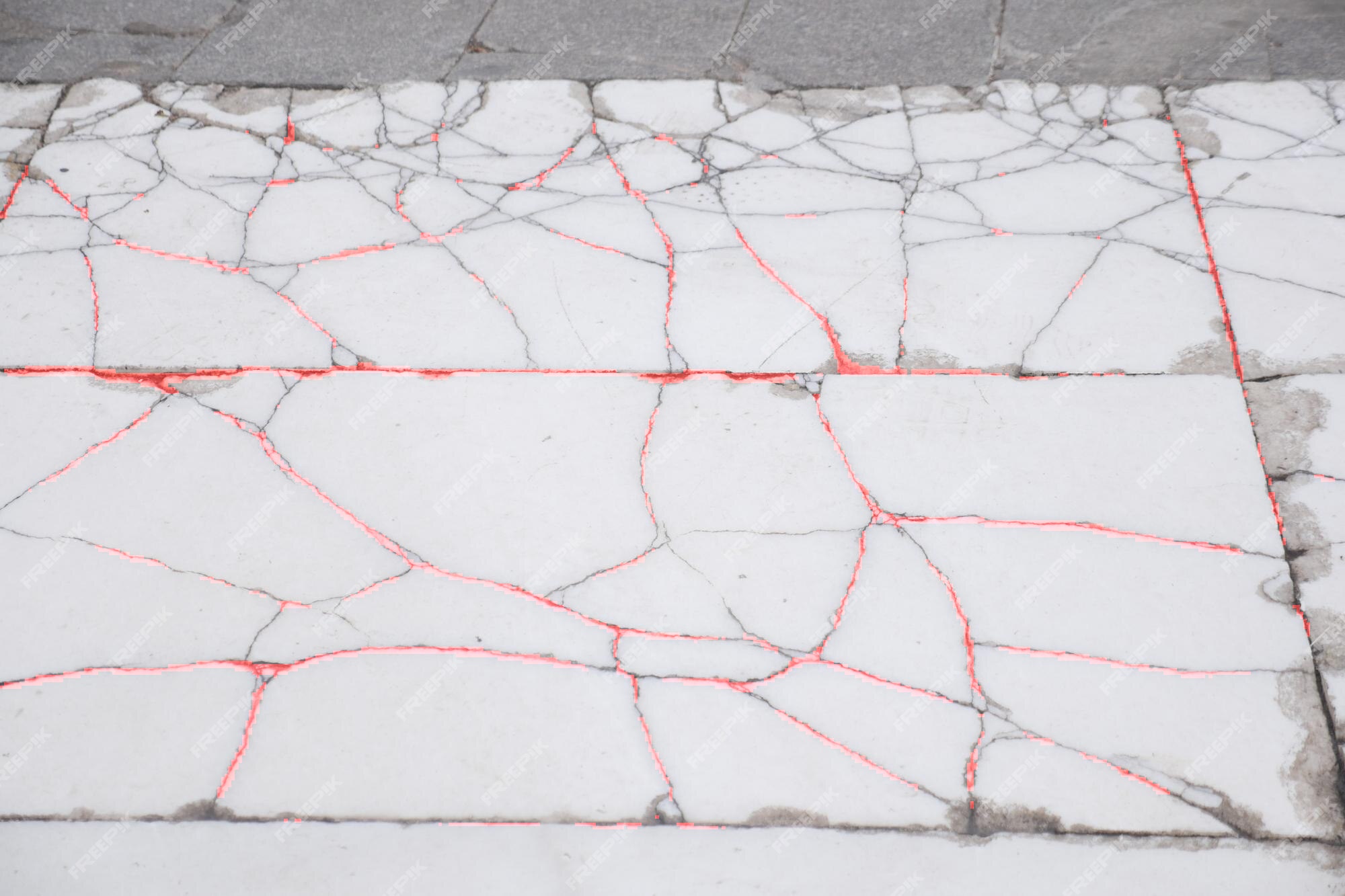}}
%\vspace{-6mm}           % SALVASPAZIO!!!
\\
\subfloat[(d) ResNet-50\label{fig:other-predictions-on-material_d}]{\includegraphics[trim={0 0 0 0},clip,width=0.325\textwidth]{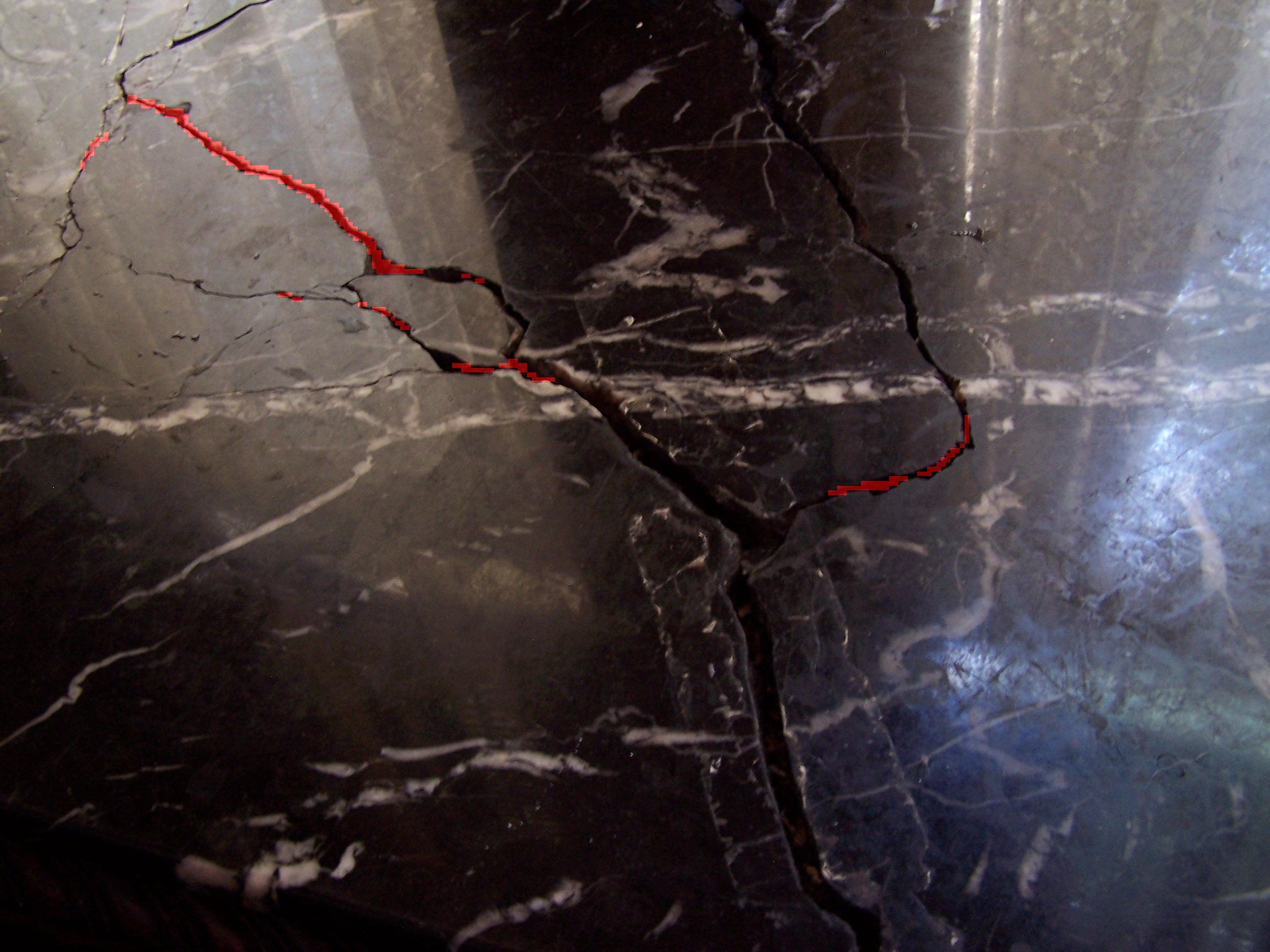}}
\hfil
\subfloat[(e) ResNet-101\label{fig:other-predictions-on-material_e}]{\includegraphics[trim={0 0 0 0},clip,width=0.325\textwidth]{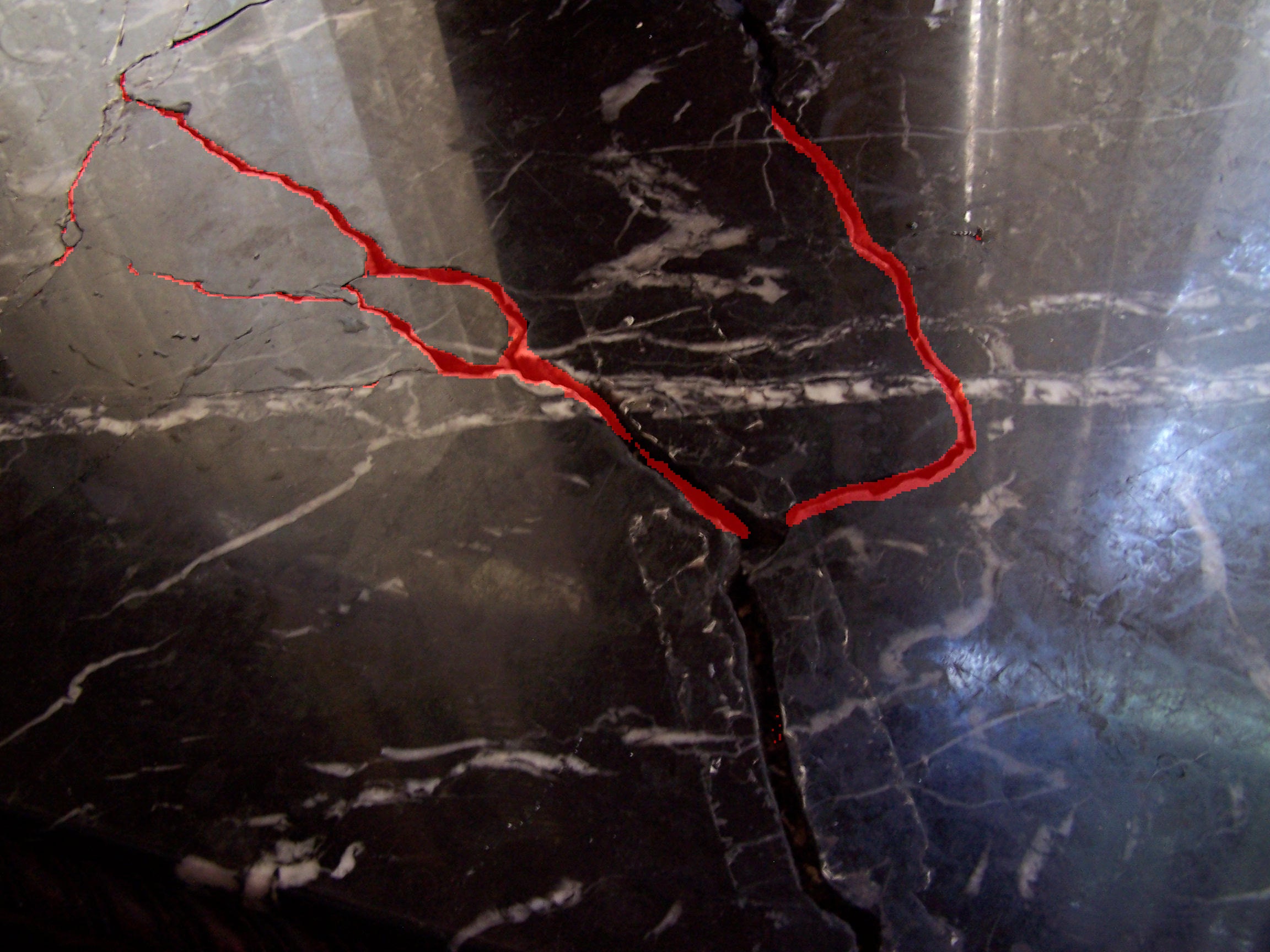}}
\hfil
\subfloat[(f) ConvNeXt V2 Base\label{fig:other-predictions-on-material_f}]{\includegraphics[trim={0 0 0 0},clip,width=0.325\textwidth]{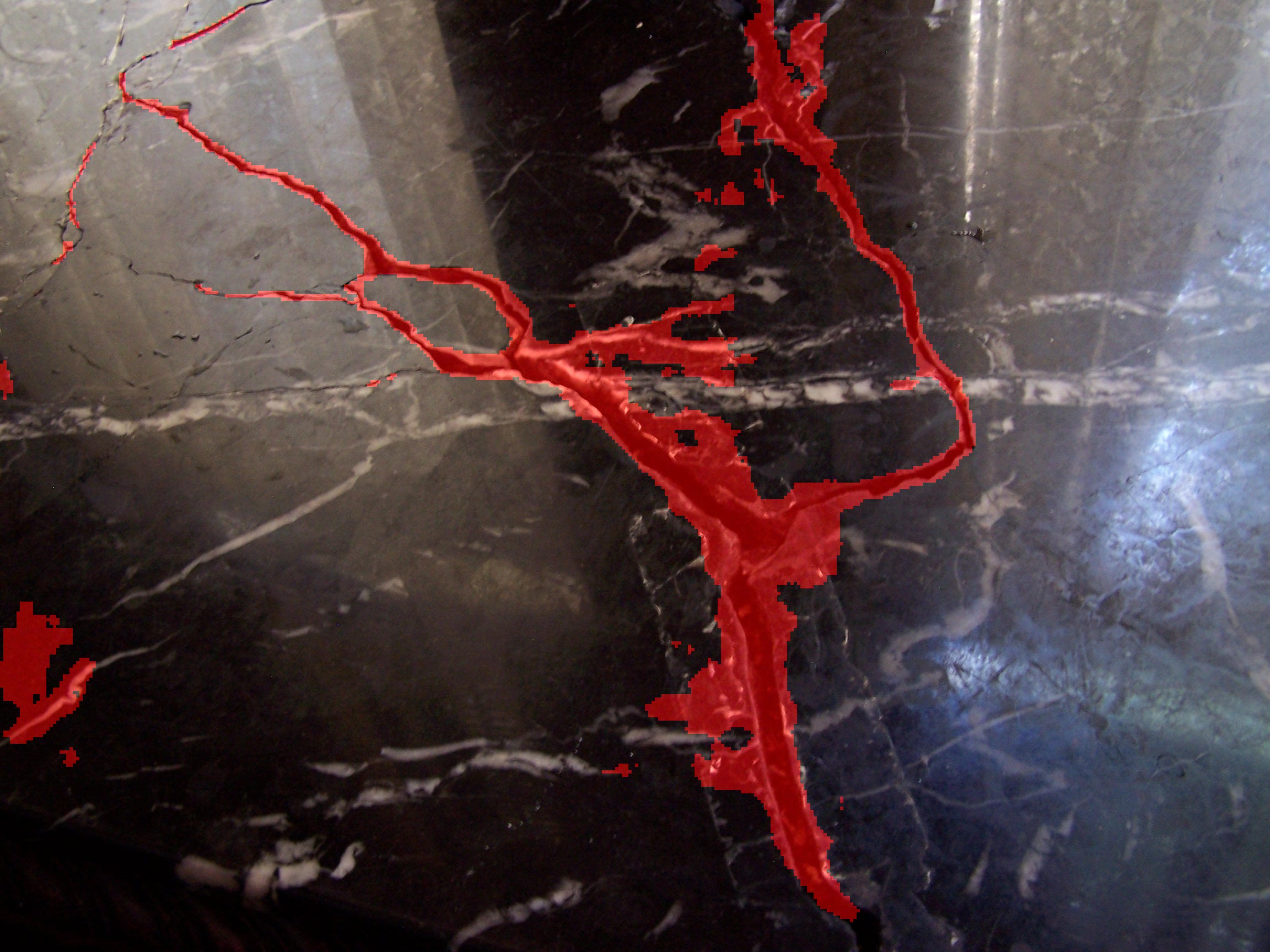}}
\vspace{-1mm}           % SALVASPAZIO!!!
\caption{The same images as in Fig. \ref{fig:ConvNext-Huge-1-predictions-on-material} processed with the other three fine-tuned models (no data augmentation regime as before).}
\label{fig:other-predictions-on-material}
\end{figure}

\begin{figure}[!thb]
\centering
\subfloat[(a)\label{fig:ConvNext-Huge-1-predictions-on-statues_a}]{\includegraphics[trim={0 0 0 0},clip,width=0.485\textwidth]{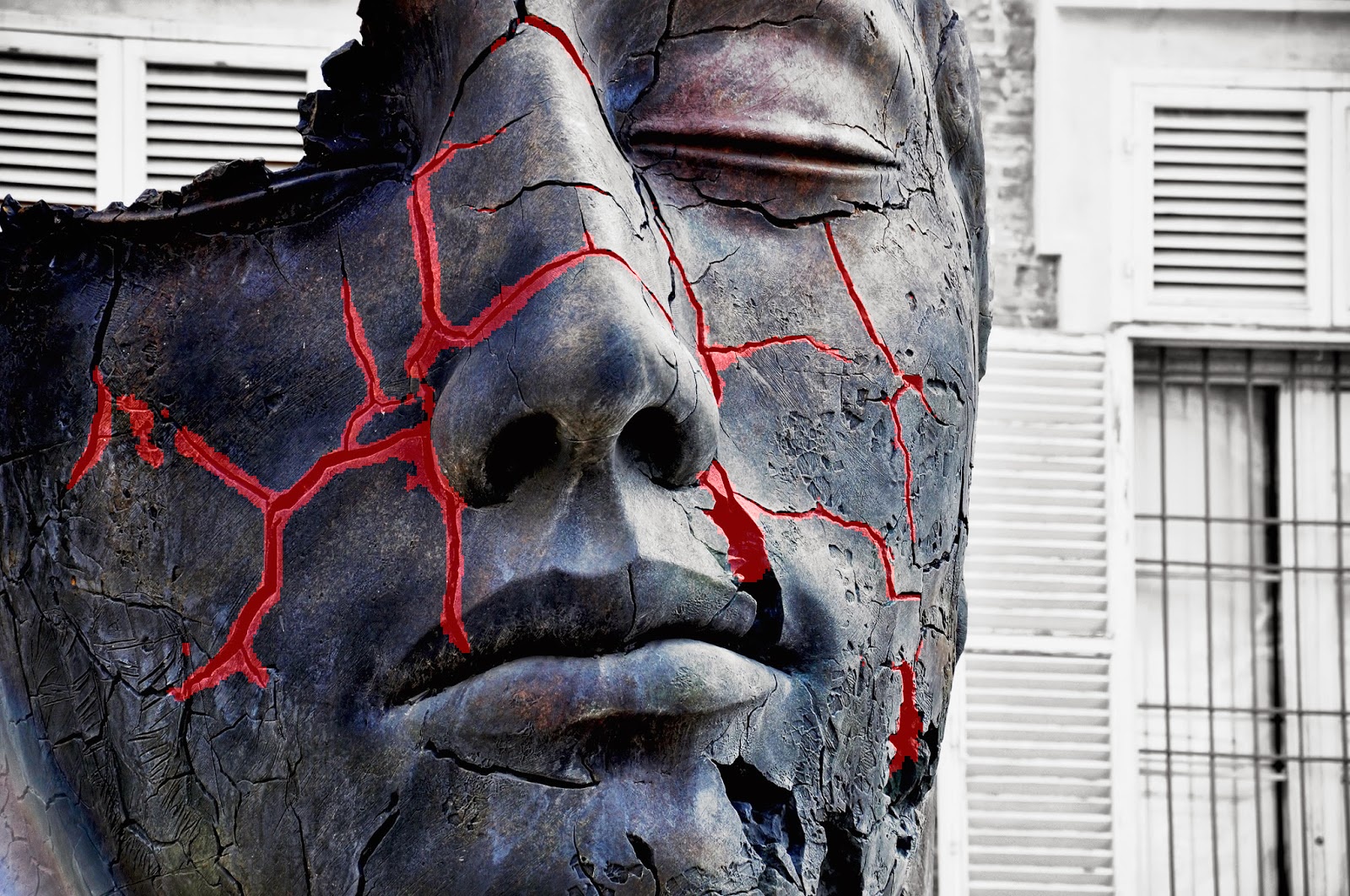}}
\hfil
\subfloat[(b)\label{fig:ConvNext-Huge-1-predictions-on-statues_b}]{\includegraphics[trim={0 0 0 0},clip,width=0.485\textwidth]{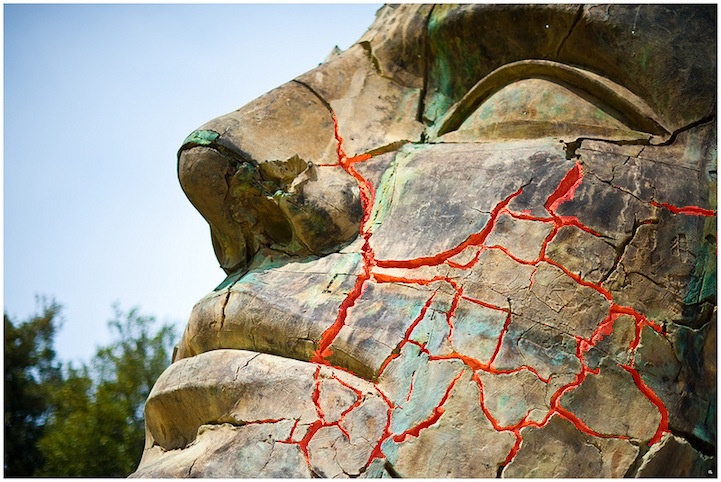}}
%\vspace{-6mm}           % SALVASPAZIO!!!
\\
\subfloat[(c)\label{fig:ConvNext-Huge-1-predictions-on-statues_c}]{\includegraphics[trim={0 0 0 0},clip,width=0.485\textwidth]{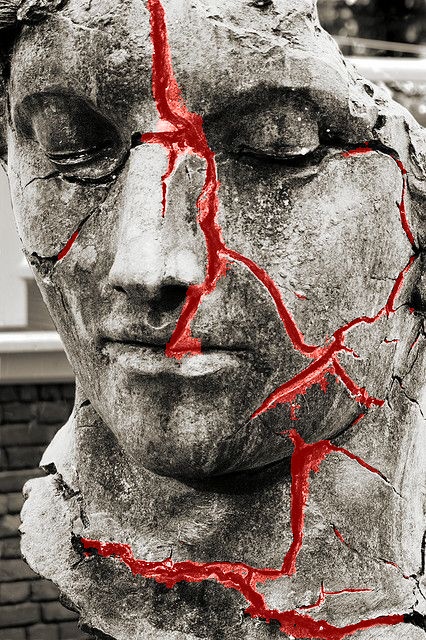}}
\hfil
\subfloat[(d)\label{fig:ConvNext-Huge-1-predictions-on-statues_d}]{\includegraphics[trim={170px 50px 50px 0},clip,width=0.485\textwidth]{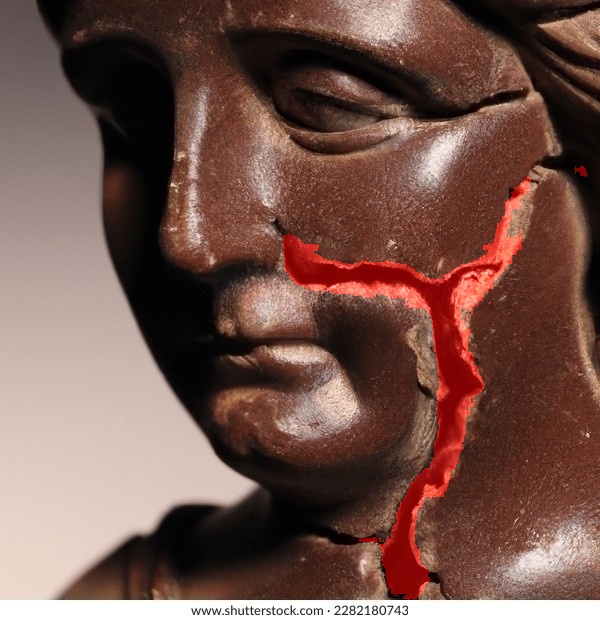}}
\vspace{-1mm}           % SALVASPAZIO!!!
\caption{Out-of-distribution predictions obtained with the \textit{ConvNeXt V2 Huge} U-Net model (no data augmentation regime) on images depicting statues (therefore quite distant from images in the training set).}
\label{fig:ConvNext-Huge-1-predictions-on-statues}
\end{figure}

\begin{figure}[!ht]
\centering
\subfloat[\label{fig:R50-45}(a) ResNet-50\label{fig:other-predictions-on-statues_a}]{\includegraphics[trim={0 0 0 0},clip,width=0.325\textwidth]{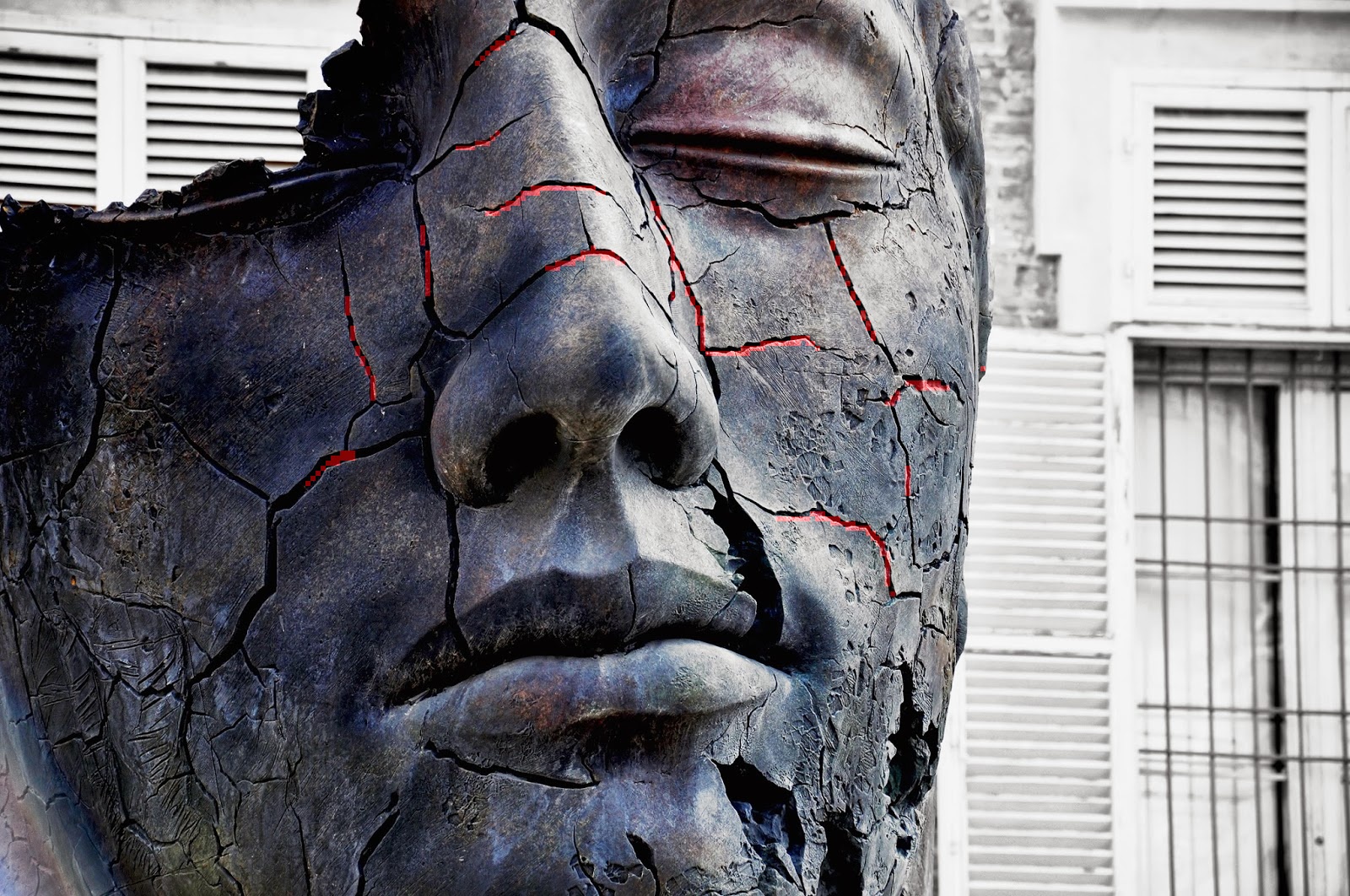}}
\hfil
\subfloat[\label{fig:R101-45}(b) ResNet-101\label{fig:other-predictions-on-statues_b}]{\includegraphics[trim={0 0 0 0},clip,width=0.325\textwidth]{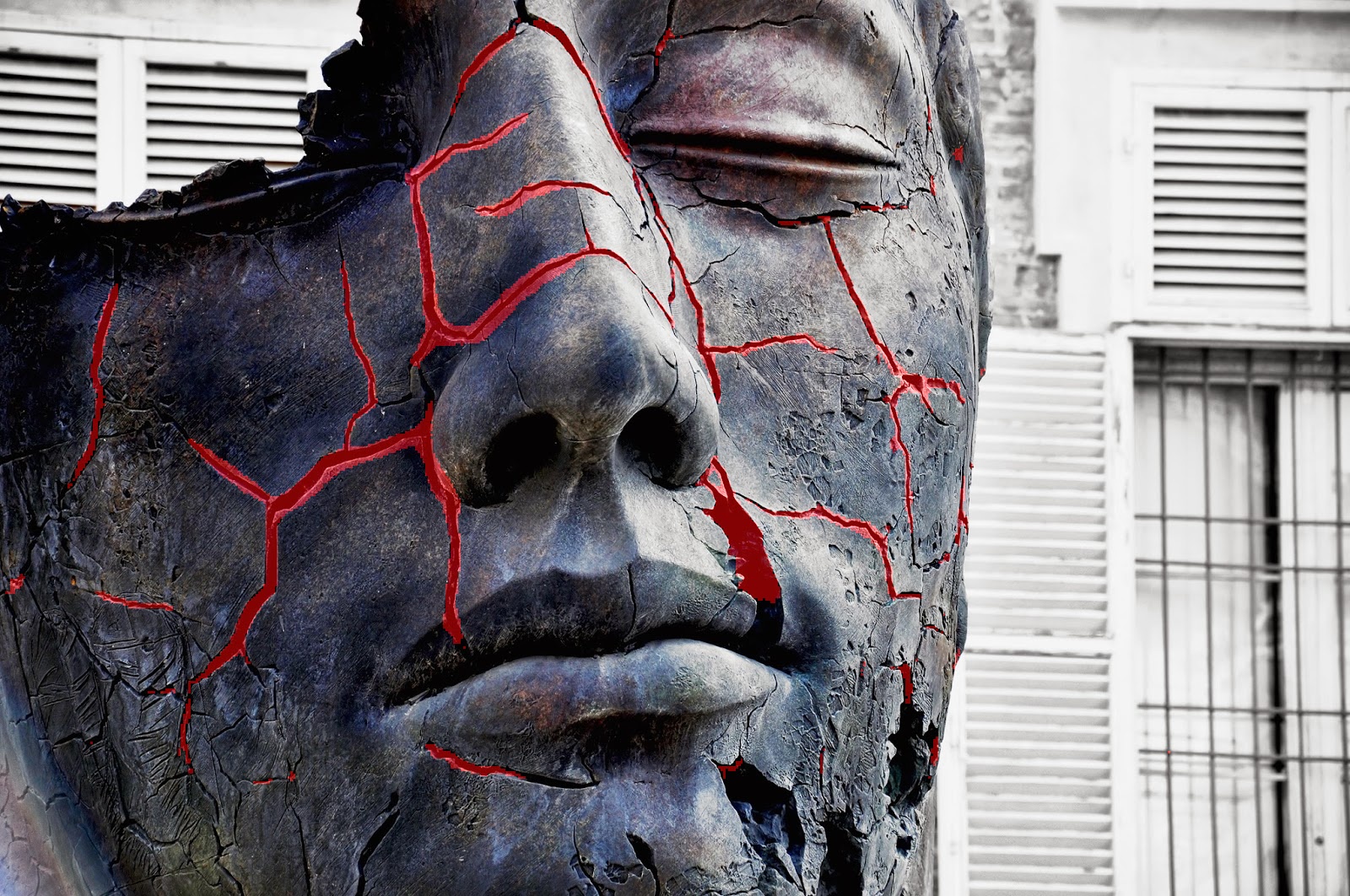}}
\hfil
\subfloat[\label{fig:CBase-45}(c) ConvNeXt V2 Base\label{fig:other-predictions-on-statues_c}]{\includegraphics[trim={0 0 0 0},clip,width=0.325\textwidth]{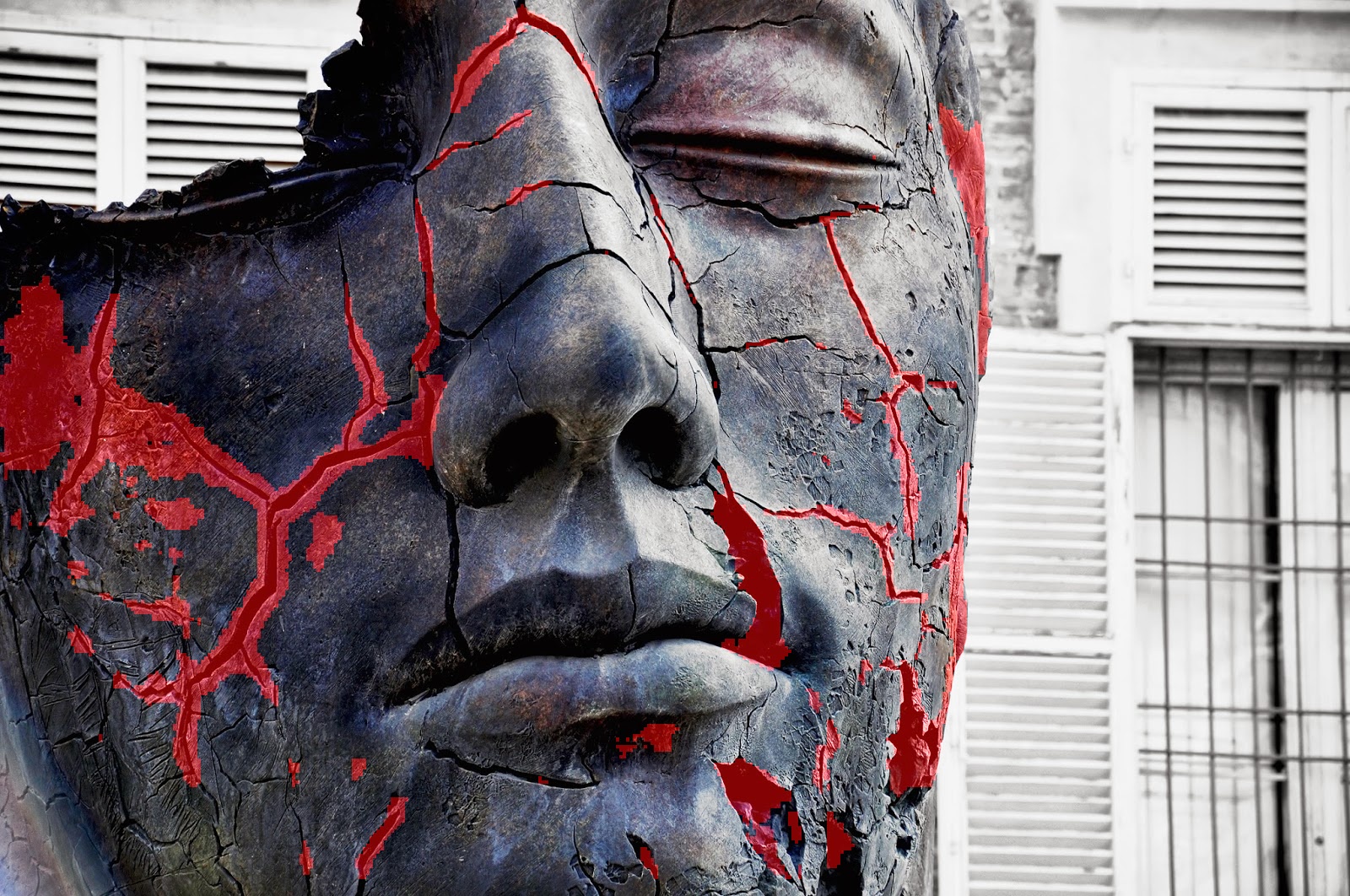}}
%\vspace{-6mm}           % SALVASPAZIO!!!
\\
\subfloat[\label{fig:R50-93}(d) ResNet-50\label{fig:other-predictions-on-statues_d}]{\includegraphics[trim={0 0 0 0},clip,width=0.325\textwidth]{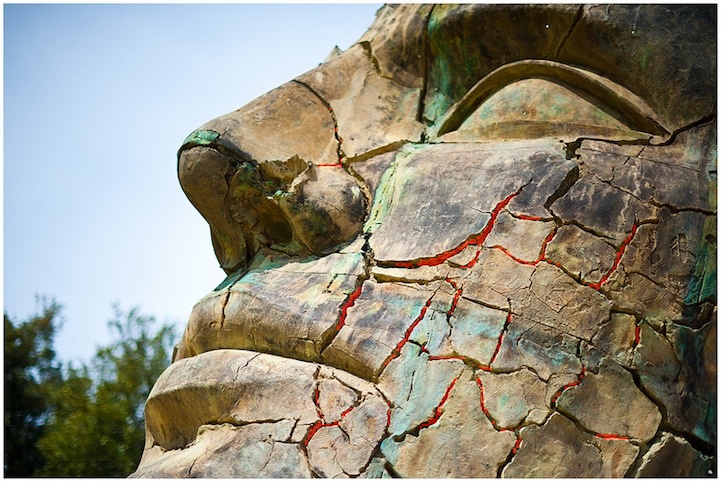}}
\hfil
\subfloat[\label{fig:R101-93}(e) ResNet-101\label{fig:other-predictions-on-statues_e}]{\includegraphics[trim={0 0 0 0},clip,width=0.325\textwidth]{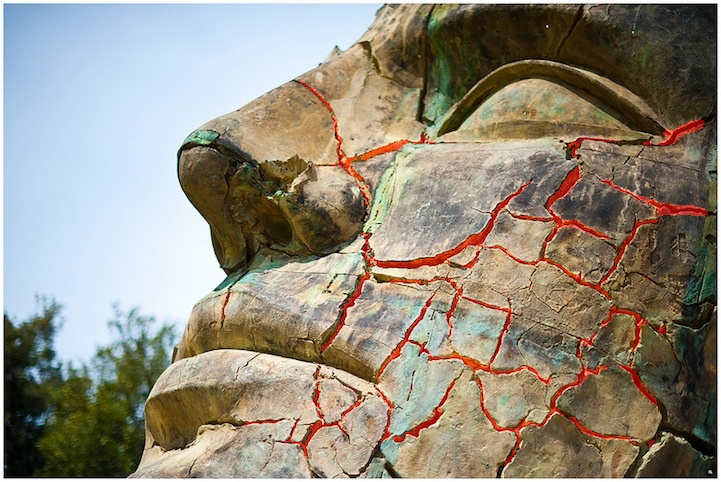}}
\hfil
\subfloat[\label{fig:CBase-93}(f) ConvNeXt V2 Base\label{fig:other-predictions-on-statues_f}]{\includegraphics[trim={0 0 0 0},clip,width=0.325\textwidth]{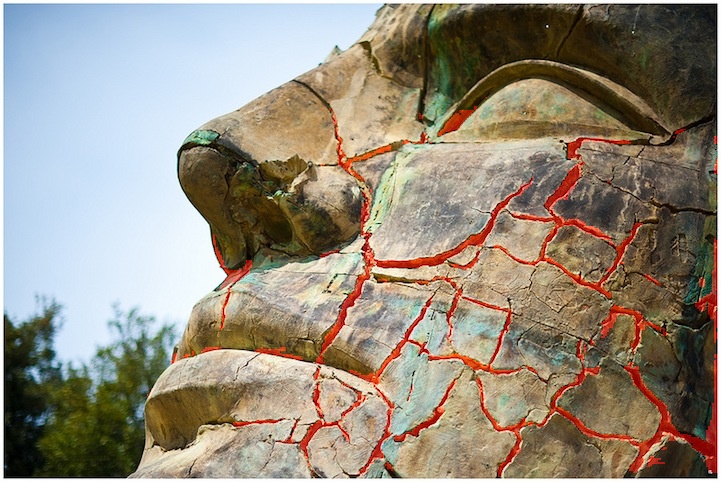}}
\\
\subfloat[\label{fig:R50-91}(g) ResNet-50\label{fig:other-predictions-on-statues_g}]{\includegraphics[trim={0 0 0 0},clip,width=0.325\textwidth]{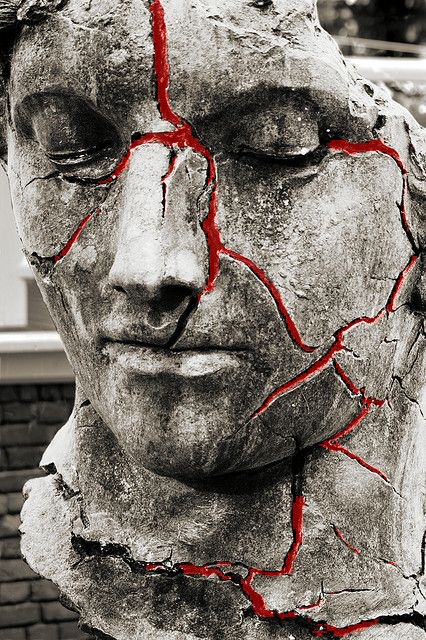}}
\hfil
\subfloat[\label{fig:R101-91}(h) ResNet-101\label{fig:other-predictions-on-statues_h}]{\includegraphics[trim={0 0 0 0},clip,width=0.325\textwidth]{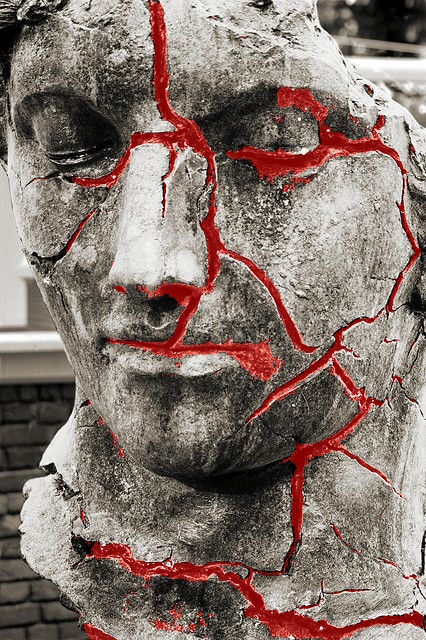}}
\hfil
\subfloat[\label{fig:CBase-91}(i) ConvNeXt V2 Base\label{fig:other-predictions-on-statues_i}]{\includegraphics[trim={0 0 0 0},clip,width=0.325\textwidth]{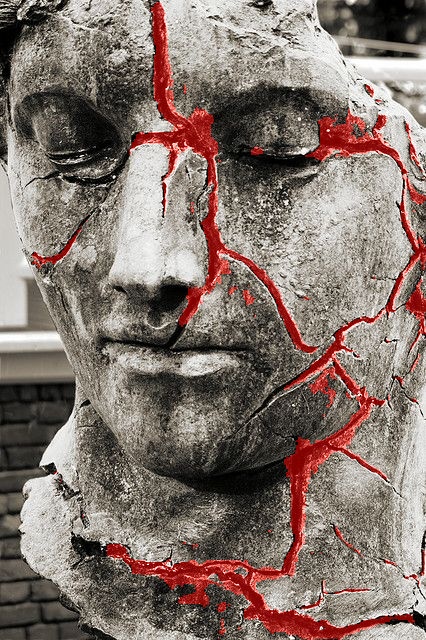}}
\\
\subfloat[\label{fig:R50-46}(j) ResNet-50\label{fig:other-predictions-on-statues_j}]{\includegraphics[trim={170px 50px 50px 0},clip,width=0.325\textwidth]{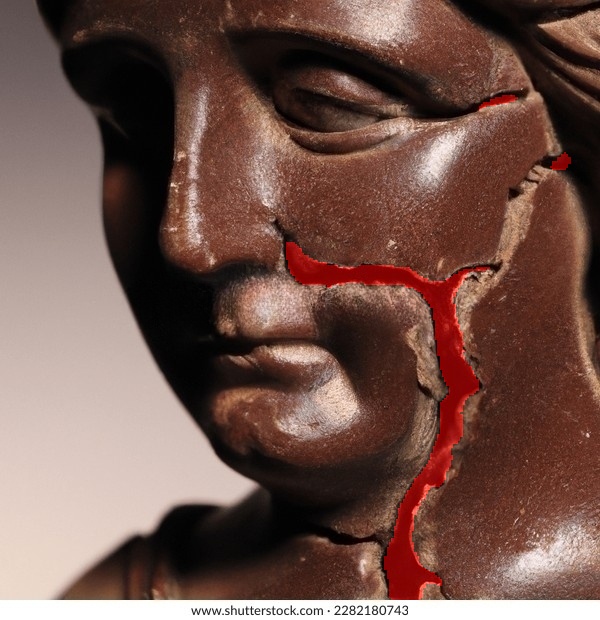}}
\hfil
\subfloat[\label{fig:R101-46}(k) ResNet-101\label{fig:other-predictions-on-statues_k}]{\includegraphics[trim={170px 50px 50px 0},clip,width=0.325\textwidth]{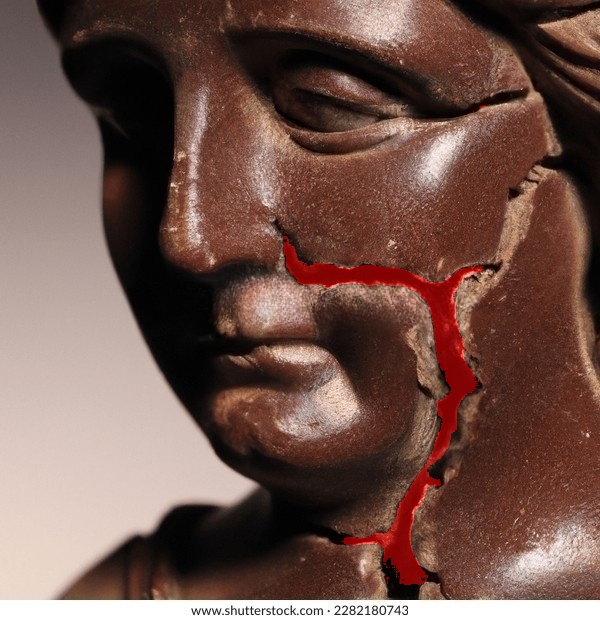}}
\hfil
\subfloat[\label{fig:CBase-46}(l) ConvNeXt V2 Base\label{fig:other-predictions-on-statues_l}]{\includegraphics[trim={170px 50px 50px 0},clip,width=0.325\textwidth]{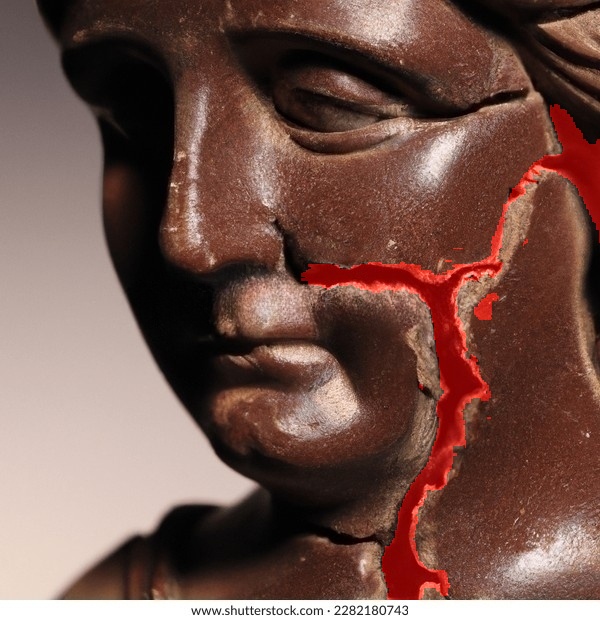}}
\vspace{-1mm}           % SALVASPAZIO!!!
\label{fig:other-predictions-on-statues}
\caption{The same images as in Fig. \ref{fig:ConvNext-Huge-1-predictions-on-statues} processed with the other three fine-tuned models (no data augmentation regime as before).}
\end{figure}

\subsection{Quantitative Evaluation on OmniCrack30k Test Set}
The quantitative results for the four U-Net architectures, trained and evaluated on the OmniCrack30k dataset, are summarized in Table \ref{tab:train_val_metrics_no_data_aug} to Table \ref{tab:test_metrics_basic_data_aug}.
%Table \ref{tab:train_val_metrics_no_data_aug}, \ref{tab:train_val_metrics_basic_data_aug}, \ref{tab:test_metrics_no_data_aug}, \ref{tab:test_metrics_basic_data_aug}.
These metrics provide insights into the models' learning efficiency and segmentation accuracy.

\subsection*{Analysis of Results}
The quantitative metrics reveal several key patterns. First, ConvNeXt V2 Huge consistently achieves the highest segmentation accuracy across all configurations (e.g., test mIoU of 0.666 without augmentation), validating its architectural superiority for artifacts crack detection. However, this comes at substantial computational cost, with 5$\times$ longer training times vs. ResNet-101. Notably, the data augmentation pipeline, as it has been conceived, \textit{reduces} the performance for most models – e.g. ConvNeXt V2 Base test mIoU drops from 0.662 to 0.647 – suggesting possible over-regularization on this dataset, eventually causing the models to "\textit{hallucinate}" cracks (false positives).

\subsection{Out-of-Distribution Qualitative Evaluation}
The qualitative evaluation on unlabeled images of real damaged and cracked statues and monuments provides crucial insights into the models' generalization capabilities to a domain not explicitly seen during training. Despite the training data (OmniCrack30k) consisting of cracks on materials like asphalt, ceramic, concrete, masonry, and steel, and not actual statues or monuments, the models demonstrate a remarkable generalization ability and are able to segment cracks in this new context. 

Specifically, as shown in Fig. \ref{fig:ConvNext-Huge-1-predictions-on-material}, the U-Net utilizing ConvNeXt V2 Huge exhibits superior performance in accurately segmenting fractures across materials with different colors and textures. In contrast, other architectures - as evidenced by Fig. \ref{fig:other-predictions-on-material} - fail to achieve comparable results. This limitation is particularly pronounced in models employing ResNet backbones. ConvNeXt V2 Base, instead, is prone to false positives segmenting black marble images, probably due to its lower image resolution and the limited exposure to such samples during training.

As illustrated in Fig. \ref{fig:ConvNext-Huge-1-predictions-on-statues}, ConvNeXt V2 Huge effectively identifies cracks in images of statues. Nevertheless, it occasionally produces false positives in the form of overflowing and imprecise boundaries.

In contrast, ResNet-based backbones exhibit some limitations, as evidenced in Fig. \ref{fig:other-predictions-on-statues}. These architectures struggle to reliably detect cracks, frequently yielding both false positives (e.g. Fig. \ref{fig:R101-91}, both the eye and the mouth are incorrectly highlighted) and false negatives (e.g. Fig. \ref{fig:R101-46}). Notably, ResNet-50 fails to detect entire cracks in several cases. While ConvNeXt V2 Base achieves reasonable segmentation performance across most statues, its predictions degrade substantially on dark-colored materials, showing pronounced false-positive artifacts.

Overall, the qualitative evaluation confirms the quantitative findings: models with more advanced and larger CNN backbones, particularly the ConvNeXt V2 variants, show better generalization capabilities for crack segmentation on real-world statues and artifacts. This is a non-trivial and non-obvious finding, as it suggests that large convolutional architectures trained on diverse crack datasets can effectively transfer their learned knowledge to new, visually distinct domains without any explicit domain-specific training, incurring in minimal false-positive rates.

%This is a significant 

\section{Conclusion}
This study presented a comparative analysis of U-Net architectures with various CNN encoders for the semantic segmentation of cracks, a critical task for cultural heritage preservation. The research leveraged the OmniCrack30k dataset for training and quantitative evaluation, acknowledging the significant challenge posed by the lack of large-scale publicly available datasets specifically annotated for the semantic segmentation of cracked statues and artifacts.

Our methodology involved training U-Net models with ResNet-50, ResNet-101, ConvNeXt V2 Base and ConvNeXt V2 Huge as encoders, at resolutions of 270px, 540px, 384px and 512px respectively. The quantitative evaluation on the OmniCrack30k test set demonstrated a clear performance hierarchy, with the ConvNeXt V2 Huge backbone achieving the highest mIoU, Dice, and Jaccard scores, followed by ConvNeXt V2 Base, ResNet-101, and ResNet-50. This highlights the benefits of increased model capacity, higher input resolution, and advanced CNN architectures like ConvNeXt V2, which benefit from masked autoencoder pre-training and for being trained on \textit{ImageNet-22k} (thus on all 14 million images).

Moreover, the out-of-distribution qualitative evaluation on images of real damaged and cracked statues and and artifacts downloaded from the web revealed a promising generalization capability. %Despite never having actually seen a single statue or monument during their training, only tiles of different cracked material from the OmniCrack30k dataset, our models are already pretty good at segmenting cracks. 
The ConvNeXt V2-based U-Nets, in particular, demonstrated superior performance in detecting fine hairline cracks, maintaining continuity, and exhibiting robustness to complex textures, materials and lighting conditions inherent in cultural heritage images. This suggests that the learned features from diverse crack patterns are highly transferable.

%Of course, there's room for improvement. While 
On the other hand, while the models already show strong generalization capabilities, fine-tuning on a dedicated, albeit small, dataset of cultural heritage cracks could further enhance their performance and reduce out-of-domain false positives.
%
%\subsection{Future Work}
%Future research should focus on several key areas to further advance automated crack detection in cultural heritage, such as

%Future research should focus primarily on the creation of large, public datasets for the semantic segmentation of statues and monuments. Aside from manual labelling, this objective can be achieved through the combination of semantic segmentation techniques for images with 3D reconstruction or synthetic generation or trying to exploiting the most recent diffusion models to achieve a "domain shift" across existing datasets (e.g. morphing datasets of persons into statues, always maintaining the same semantic boundaries).

Future research should focus primarily on the creation of large-scale, publicly available datasets for statue and monument segmentation. Beyond manual annotation, this objective can be achieved combining image segmentation with 3D reconstruction, with synthetic 3D data generation or leveraging diffusion models for domain adaptation (e.g. adapting human-centric datasets to the statue domain while preserving semantic boundaries).
%\textcolor{blue}{
Moreover, future efforts should involve the technical expertise of conservators and archaeologists, potentially for labeling future datasets, providing qualitative assessment of model performance, and, most importantly, identifying underrepresented artifact types (e.g. beyond statues and monuments) critical to cultural heritage preservation.%}
%; the adoption of domain adaption techniques; the combination of semantic segmentation techniques for images with 3D reconstruction, and development of methods that provide intuitive insights into the model prediction. 
%\begin{itemize}
%\item \textbf{Curated Cultural Heritage Crack Datasets:} While our qualitative evaluation is valuable, the creation of even small, expertly annotated real-world datasets of cracks on statues and monuments would enable more rigorous quantitative evaluation and targeted model development.
%\item \textbf{Advanced Domain Adaptation:} Exploring more sophisticated domain adaptation or few-shot learning techniques could further bridge the gap between general crack datasets and the specific visual characteristics of cultural heritage materials.
%\item \textbf{Integration with 3D Reconstruction:} Combining 2D semantic segmentation with 3D reconstruction techniques could provide a more comprehensive understanding of crack depth and structural impact, offering invaluable insights for conservators.
%\item \textbf{Real-time Deployment and Edge Computing:} Optimizing these models for real-time inference on portable devices or drones would facilitate efficient on-site inspections in the field.
%\item \textbf{Explainable AI for Conservation:} Developing methods to provide conservators with interpretable insights into model predictions (e.g., confidence maps, feature visualizations) could increase trust and adoption.
%\end{itemize}
In conclusion, this research represents a significant step towards leveraging advanced deep learning for the automated preservation of our invaluable global cultural heritage.

\section*{Acknowledgements}
%\textcolor{blue}{
This work was carried out within the framework of the Italian Ministry of Business and Made in Italy (MIMIT) project, House of Emerging Technologies Genoa (CTEGE): Digital factory for culture.%}

%\textcolor{blue}{
This work was also partially funded within the framework of the activities of the National Recovery and Resilience Plan (NRRP) M4C2 Inv. 1.4 -- CN MOST -- Sustainable Mobility Center, Spoke 7, whose financial support is gratefully acknowledged.%}

 \bibliographystyle{unsrt} %splncs04
 \bibliography{bibliography}
%Author, F.: Article title. Journal \textbf{2}(5), 99--110 (2016)

%Author, F., Author, S.: Title of a proceedings paper. In: Editor,
%F., Editor, S. (eds.) CONFERENCE 2016, LNCS, vol. 9999, pp. 1--13.
%Springer, Heidelberg (2016). \doi{10.10007/1234567890}
%Author, F., Author, S., Author, T.: Book title. 2nd edn. Publisher,

%Author, A.-B.: Contribution title. In: 9th International Proceedings
%on Proceedings, pp. 1--2. Publisher, Location (2010)
%LNCS Homepage, \url{http://www.springer.com/lncs}. Last accessed 4
%\end{thebibliography}

%\pagebreak
\newpage
\appendix
\section{Appendix}
This appendix presents an ablation study evaluating the impact of individual data augmentation techniques on the performance of our smallest architecture, a \textit{U-Net} with a \textit{ResNet-50} encoder trained with images at \textit{270px} resolution. Each model was trained on the OmniCrack30k dataset using a single augmentation transform, with validation and test metrics reported in Tables \ref{tab:ablation_validation_metrics} and \ref{tab:ablation_test_metrics}, respectively. Key observations are summarized below.

\vspace{-5mm}           % SALVASPAZIO!!!
\begin{table}[!h]
\centering
\begin{tabular}{|>{\color{black}}l|>{\color{black}}c|>{\color{black}}c|>{\color{black}}c|>{\color{black}}c|}
\hline
\textbf{Augmentation} & \textbf{Valid Loss} & \textbf{mIoU} & \textbf{Dice} & \textbf{Jaccard} \\
\hline
% Sorted by Dice Multi
Transpose & 0.025 & 0.643 & 0.854 & 0.772 \\
CLAHE & 0.026 & 0.643 & 0.851 & 0.769 \\
GridDistortion & 0.026 & 0.646 & 0.849 & 0.766 \\
ElasticTransform & 0.027 & 0.641 & 0.845 & 0.761 \\
RandomRotate90 & 0.028 & 0.633 & 0.843 & 0.759 \\
OpticalDistortion & 0.027 & 0.633 & 0.841 & 0.756 \\
Blur & 0.028 & 0.635 & 0.839 & 0.755 \\
HorizontalFlip & 0.028 & 0.631 & 0.837 & 0.752 \\
ShiftScaleRotate & 0.028 & 0.628 & 0.834 & 0.749 \\
HueSaturationValue & 0.033 & 0.625 & 0.821 & 0.734 \\
\hline
\end{tabular}
\vspace{3mm}           % CREASPAZIO!!!
\caption{Ablation study on data augmentation techniques: validation metrics sorted by Dice score for ten different \textit{ResNet-50 @270px} models trained with a single \textit{Albumentation} transform type each.}
\label{tab:ablation_validation_metrics}
\end{table}

\vspace{-16mm}           % SALVASPAZIO!!!
\begin{table}[!h]
\centering
\begin{tabular}{|>{\color{black}}l|>{\color{black}}c|>{\color{black}}c|>{\color{black}}c|>{\color{black}}c|}
\hline
\textbf{Augmentation} & \textbf{Test Loss} & \textbf{mIoU} & \textbf{Dice} & \textbf{Jaccard} \\
\hline
% Sorted by Dice Multi
Transpose & 0.023 & 0.675 & 0.866 & 0.787 \\
GridDistortion & 0.024 & 0.672 & 0.863 & 0.783 \\
CLAHE & 0.025 & 0.668 & 0.861 & 0.780 \\
ElasticTransform & 0.025 & 0.669 & 0.856 & 0.775 \\
RandomRotate90 & 0.026 & 0.660 & 0.854 & 0.772 \\
Blur & 0.026 & 0.661 & 0.854 & 0.772 \\
OpticalDistortion & 0.025 & 0.658 & 0.852 & 0.770 \\
HorizontalFlip & 0.025 & 0.655 & 0.850 & 0.768 \\
ShiftScaleRotate & 0.026 & 0.655 & 0.842 & 0.758 \\
HueSaturationValue & 0.034 & 0.650 & 0.834 & 0.749 \\
\hline
\end{tabular}
\vspace{3mm}           % CREASPAZIO!!!
\caption{Ablation study on data augmentation techniques: test metrics sorted by Dice scorefor ten different \textit{ResNet-50 @270px} models trained with a single \textit{Albumentation} transform type each.}
\label{tab:ablation_test_metrics}
\end{table}

%\textcolor{blue}{
The ablation study shows that top-performing augmentations, selected by their \textit{Dice} score, are \textit{Transpose}, \textit{CLAHE}, and \textit{GridDistortion} with \textit{Dice} scores ranging from 0.854 to 0.849 on the validation set and 0.866–0.861 on the test set\footnote{%\textcolor{blue}{
Slightly higher \textit{Dice} score values in the test set compared to those in the validation set are consistent with what was observed in Tables \ref{tab:train_val_metrics_no_data_aug} and \ref{tab:test_metrics_no_data_aug}, probably due to the intrinsic conformation of the dataset.}
%}
. These techniques likely enhance crack detection for the following reasons:
\begin{itemize}
    \item \textbf{Transpose} preserves structural patterns while altering orientation in 90° steps, improving robustness to spatial variations.
    \item \textbf{CLAHE} (Contrast Limited Adaptive Histogram Equalization) enhances local contrast in low-light regions, amplifying subtle crack features without introducing further noise.
    \item \textbf{GridDistortion} simulates natural surface deformations (e.g., material warping), aiding generalization to unseen images.
\end{itemize}
%}

%\textcolor{blue}{
On the other hand, the most detrimental augmentations are \textit{HueSaturationValue} and \textit{ShiftScaleRotate} with Dice scores ranging from 0.821 to 0.834 on the validation set and 0.834-0.842 on the test set. Potential reasons include:
\begin{itemize}
    \item \textbf{HueSaturationValue} distorts color channels, obscuring grayscale crack features critical for detection.
    \item \textbf{ShiftScaleRotate} performs translations, scaling and rotations, the combination of which can erase the finest cracks from the image or make them too blurred (due to rotation, in fact, \textit{Blur}, understandably, has similar impacts on segmentation).
\end{itemize}
%}

\end{document}